\documentclass[a4paper,11pt]{article}

\usepackage[margin=1in]{geometry}
\usepackage{authblk}

\usepackage[utf8]{inputenc} % allow utf-8 input
\usepackage[T1]{fontenc}    % use 8-bit T1 fonts
\usepackage{hyperref}       % hyperlinks
\usepackage{url}            % simple URL typesetting
\usepackage{booktabs}       % professional-quality tables
\usepackage{amsfonts}       % blackboard math symbols
\usepackage{nicefrac}       % compact symbols for 1/2, etc.
\usepackage{microtype}      % microtypography
\usepackage{xcolor}         % colors

\usepackage{microtype}
\usepackage{graphicx}
\usepackage{subcaption}
\usepackage{booktabs} % for professional tables
\usepackage{hyperref}

\usepackage{natbib}
\renewcommand{\cite}[1]{\citep{#1}}

\usepackage{amsmath}
\usepackage{amssymb}
\usepackage{mathtools}
\usepackage{amsthm}

\usepackage{setspace}
\usepackage{etoolbox}

\usepackage[boxed,vlined]{algorithm2e}
    \SetInd{0.3em}{0.7em}
    \makeatletter
    \patchcmd{\@algocf@start}{-1.5em}{0pt}{}{}
    \makeatother
    \SetCommentSty{graycomment}

    \SetKwProg{Fn}{}{:}{end}

    \SetAlCapSkip{.8\baselineskip}
\usepackage[capitalize,noabbrev]{cleveref}
\Crefname{equation}{Eq.}{Eqs.}

\usepackage{enumitem}
\usepackage{pgfplots}
\usepackage{pgfplotstable}

\usepackage{tikz}
\usetikzlibrary{shapes.geometric, shapes.misc, automata, positioning, arrows, arrows.meta, decorations.markings}
\usetikzlibrary{calc}
\usepackage{pgfmath}

\BeforeBeginEnvironment{tikzpicture}{\vspace{3mm}}
\AfterEndEnvironment{tikzpicture}{\vspace{1mm}}

\usepackage{wrapfig}
\usepackage{caption}
\usepackage{thm-restate}

\usepackage{framed}

\usepackage{ifthen}
\newboolean{isappendix}
\setboolean{isappendix}{false}
\newcommand{\hideinrestate}[1]{\ifthenelse{\boolean{isappendix}}{}{#1}}

\newtheorem{thm}{_} % fix

\newtheorem{corollary}[thm]{Corollary}

\newtheorem{claim}{Claim}

\theoremstyle{definition}
\newtheorem{definition}{Definition}

\newcommand{\barpi}{\bar{\pi}}

\newcommand{\true}{\mathit{true}}
\newcommand{\false}{\mathit{false}}
\newcommand{\VV}{\vec{V}}
\newcommand{\vv}{\mathbf{v}}
\newcommand{\xcalV}{\widehat{\calV}}
\newcommand{\xcalB}{\widehat{\calB}}
\newcommand{\xPhi}{\widehat{\Phi}}
 
\newcommand{\conv}{\operatorname{Conv}}

\newcommand{\neigh}{\operatorname{Neigh}}
\newcommand{\gridhull}{\operatorname{GridHull}}
\newcommand{\bellman}{\operatorname{Bellman}}
\newcommand{\onward}{\operatorname{Onward}}
\newcommand{\Expand}{\textnormal{\texttt{Expand}}}
\newcommand{\Achieve}{\textnormal{\texttt{Achieve}}}
\newcommand{\ActProbs}{\textnormal{\texttt{ActProbs}}}
\newcommand{\VSI}{\textnormal{\texttt{VSI}}}

\renewcommand{\emptyset}{\varnothing}

\newcommand{\vsirepourl}{{\color{blue}\url{https://github.com/jrgan/VSI}}}

\usepackage{bm}
\newcommand{\nn}{\mathbf{n}}

\newcommand{\ww}{\mathbf{w}}
\newcommand{\xx}{\mathbf{x}}
\newcommand{\yy}{\mathbf{y}}
\newcommand{\zz}{\mathbf{z}}

\newcommand{\calB}{\mathcal{B}}

\newcommand{\calV}{\mathcal{V}}

\DeclareMathOperator*{\argmax}{arg\,max}

\DeclareMathOperator{\E}{\mathbb{E}}
\newcommand{\given}{{\,|\,}}

\title{Subprocess-Constrained Markov Decision Processes}

\author[1]{Jiarui Gan}
\author[2]{Debmalya Mandal}

\affil[1]{University of Oxford}
\affil[2]{University of Warwick}

\date{}

\begin{document}

\maketitle

\begin{abstract}
Constrained Markov decision processes (CMDPs) typically bound expected cumulative cost from an initial state.
Such a guarantee can be insufficient when a policy must remain feasible after every history it may encounter.
We introduce \emph{subprocess-constrained MDPs} (S-CMDPs), which require the expected discounted cost of each continuation, conditional on any reachable history, to satisfy a bound determined by the current state.
This stronger requirement changes the structure and complexity of the problem: we show that stationary policies can be strictly suboptimal, and finding even an approximately optimal stationary policy is NP-hard.
We adapt a recently introduced value-set iteration framework to show that history-dependent policies surprisingly overcome both obstacles.
Our polynomial-time algorithm represents such a policy implicitly: its expected reward is within $\epsilon$ of the optimal feasible value, it violates each subprocess cost bound by at most $\delta$, and its action distribution can be computed in polynomial time for any given history.
Code is available at \vsirepourl.
\end{abstract}

\section{Introduction}

Markov decision processes (MDPs) are a standard model for sequential decision-making under uncertainty.
At each step, an agent observes the state of the environment, chooses an action, and receives a reward as the environment evolves stochastically.
In many domains, however, decision-making is subject to requirements that are {\em not} captured by reward alone: energy consumption must stay within a budget, risk must remain below a safety threshold, and violations of service-level targets must be limited.
Constrained Markov decision processes (CMDPs)~\cite{altman2021constrained} address such requirements by augmenting an MDP with a {\em cost} function and imposing a constraint, typically on the expected cumulative cost over the entire trajectory.
This global, expectation-based formulation has a long history in stochastic control and operations research and remains central to modern safe reinforcement learning~\cite{garcia2015comprehensive,gu2024review}.

Yet, in safety- and reliability-critical applications, the classical CMDP viewpoint can be misaligned with how systems are deployed and assessed. Standard constraints are evaluated only from the initial state, providing guarantees tied to the beginning of the trajectory. In many settings, the continuation from time step $t$ should satisfy the same requirements as a process starting anew. It is therefore natural to require the constraints imposed on the full process to hold for \emph{every subprocess} following a reachable history.
Without such subprocess constraints, a policy may satisfy a global expected-cost budget while behaving unacceptably after certain histories. For example, low early costs may offset high future costs under the global constraint, or high-cost trajectories may be averaged against low-cost ones.

Such concerns arise naturally in practice. For instance, when an autonomous robot reaches a crowded region or a narrow passage near humans, safety in expectation from the initial deployment state is not enough; from the newly reached configuration, the controller should still keep future collision risk below an acceptable threshold~\cite{fisac2018probabilistically,zhu2019chance}. Similarly, in online services governed by service-level objectives (SLOs), reliability is monitored continuously through service-level indicators and error budgets. Exhausting an error budget can trigger responses such as reliability work or deployment freezes~\cite{srebook2016,googleSREworkbook2018}. It is unacceptable to enter a congested state from which expected future latency or availability violations exceed acceptable thresholds, even if the initial expected violations remain within budget.
In these settings, constraints that apply only at initialization can be too weak.

\subsection{Our Work and Contributions}

To address this limitation, we introduce \emph{subprocess-constrained MDPs} (S-CMDPs). Informally, S-CMDPs require policies to remain feasible under a ``restart at time $t$'' interpretation after every reachable history, capturing a subprocess-level notion of safety and reliability not enforced by classical CMDPs. Although this may appear to be a modest change in formulation, it fundamentally changes the problem: standard CMDP solution methods do not transfer directly.

We study the algorithmic aspects of S-CMDPs and establish the following foundational results:
\begin{itemize}
    \item 
    We identify {\bf a fundamental computational separation} between S-CMDPs and standard CMDPs. Stationary policies, which suffice for optimality and can be optimized efficiently in classical MDPs and CMDPs, can be strictly suboptimal in S-CMDPs. More importantly, even if we accept this loss of optimality and restrict attention to stationary policies, finding the best policy in that class, even just approximately, is NP-hard.

    \item 
    We show that history-dependent policies, which can be necessary for optimality, can be handled efficiently using {\bf value-set iteration (VSI)}.
    This is surprising because representing a history-dependent policy explicitly requires space exponential in the horizon.
    In S-CMDPs, allowing history dependence is key to establishing {\bf tractability}.
    Our approach leverages an implicit representation of history-dependent policies: rather than explicitly specifying action distributions, the policy is given by an algorithm that efficiently computes its action distribution for any given history.

    \item 
    VSI extends classical value iteration to a {\em set-valued} dynamic programming framework. Unlike classical value iteration, the associated set-valued operator is generally not a contraction, so convergence requires a different analysis. We characterize the solution of an S-CMDP as a {\bf fixed point} of this operator and develop a {\bf bicriteria approximation scheme} that computes a fixed point of an approximate operator in polynomial time. This yields an $(\epsilon,\delta)$-optimal history-dependent policy, in contrast to the NP-hardness of computing such a stationary policy.
\end{itemize}

Our results establish S-CMDPs as a distinct extension of CMDPs that enforces a stronger notion of safety while remaining tractable through our VSI framework.

\subsection{Related Work}

Our work builds on several strands of CMDP and safe RL research, while differing from them in its treatment of onward feasibility.
Classical CMDPs constrain expected cumulative cost from the initial state and admit strong structural results and solution methods, including linear programming and Lagrangian optimization~\cite{altman1998constrained,altman2021constrained}. 
Much of modern safe RL inherits this formulation; for example, constrained policy optimization (CPO)~\cite{achiam2017constrained} targets global constraints while providing guarantees about constraint satisfaction.
Surveys of safe RL~\cite{garcia2015comprehensive,gu2024review} similarly center on initial-state expectation constraints and closely related risk criteria. 
Our S-CMDP model imposes a stronger feasibility requirement: onward feasibility after every reachable history, rather than only feasibility in expectation from the start.

Several lines of work in safe RL and constrained MDPs strengthen safety requirements at individual times or states.
For example, peak-constrained CMDPs~\citep{BAG23} require constraints to hold at every time step with probability one; in such formulations, the constraint can often be viewed as inducing a safe action set at each state-time pair. 
Related safe RL formulations impose instantaneous hard constraints that rule out unsafe actions or unsafe states during learning~\citep{amani2021safe,shi2023near,wei2024safe}. 
A distinct line of work studies instantaneous constraints in expectation and develops Lagrangian or augmented-Lagrangian methods for enforcing time-indexed constraint violations~\citep{li2021augmented}. 
Recent work on state-wise constrained policy optimization also imposes constraints at individual states~\citep{zhaostate2024}, including variants that guide learning by requiring visited states to remain within a safe state set~\citep{sun25a}.

These formulations are complementary in motivation, but they primarily constrain local quantities: immediate action safety, next- or visited-state safety, worst-case state-wise violations, or expected instantaneous violations.
By contrast, S-CMDPs constrain the entire discounted continuation from each reachable history, which spans multiple or even infinite time steps rather than only the next action, transition, or visited state.

Closer to our model is the recent work on anytime-constrained MDPs~\citep{mcmahan2024anytime}. 
Anytime-constrained MDPs also recognize that feasibility from the initial state alone may be insufficient, and instead require the agent to avoid violating its budget at any time. 
\citet{mcmahan2024anytime} show, among other results, that Markovian policies are generally insufficient unless augmented with additional information, such as cumulative cost. 
Our S-CMDP formulation shares this motivation but bounds the expected onward cost of every subprocess rather than the accumulated cost along every prefix.
This distinction leads to different algorithmic techniques. 
Anytime-constrained MDPs motivate state augmentation by cumulative cost or related bookkeeping variables~\citep{mcmahan2024anytime}. 
In contrast, we use value-set reasoning over attainable subprocess values, which enables a direct characterization of feasible continuations and a polynomial-time procedure for computing near-optimal history-dependent policies.

On the algorithmic side, value-set iteration has precedents in game theory, where similar set-valued dynamic programming methods have been used to compute equilibria in sequential games~\cite{murray2007finding,dermed2009solving,kitti2016subgame,gan2023sequentialv2,gan2025valueset}. 
In these settings, players' incentive constraints are defined in terms of expected future payoffs and must be satisfied in every subgame, mirroring the subprocess constraints in our model. 
The recent work of \citet{gan2025valueset} provides a detailed algorithmic study of value-set iteration for computing extensive-form correlated equilibria in Markov games. 
Our work adapts this framework to S-CMDPs, a setting that we believe admits an even broader range of applications in planning, control, and RL.
At a technical level, our VSI framework rests on a fixed point of a monotone operator.
The computation of Tarski fixed points has recently attracted growing attention in theoretical computer science~\cite{etessami2019tarski,fearnley2022faster,chen2022improved,batziou2025monotone}.

\section{The S-CMDP Model}
\label{sc:prelim}

An S-CMDP is given by a tuple $(S, A, s_0, p, r, c, \theta, \gamma)$.
The components other than $c$ and $\theta$ define a standard discounted MDP:
$S$ is a finite state space, with initial state $s_0 \in S$;
$A$ is a finite action space;
$p: S \times A \to \Delta(S)$ is the transition kernel;
$r: S \times A \to \mathbb{R}$ is the reward function;
and $\gamma \in (0,1)$ is the discount factor.
The additional components $c$ and $\theta$ specify a cost function $c : S \times A \to \mathbb{R}$ and a state-wise threshold $\theta(s) \in \mathbb{R}$ for each $s \in S$.
Without loss of generality, we assume that rewards and costs lie in $[-1,1]$.

Let a policy $\pi$ assign an action distribution $\pi(h) \in \Delta(A)$ to each finite history $h = (s_0, a_0, \dots, s_t)$.
For a reachable history $h$, we define the onward reward value as
\[
\textstyle
V_r^\pi(h) = \E_\pi \left[ \sum_{t'=t}^\infty \gamma^{t'-t} \cdot r(s_{t'}, a_{t'}) \,\middle|\, h \right],
\]
where $\E_\pi[\cdot\mid h]$ denotes expectation conditional on $h$ under $\pi$, and $\mathbb{P}_\pi(h)$ denotes the probability of observing $h$.
Similarly, $V_c^\pi(h)$ denotes the onward discounted cost, defined by replacing $r$ with $c$.

Our objective is to maximize expected discounted reward while bounding the onward discounted cost after every reachable history:
\begin{align}
\hspace{-5mm}
\max_{\pi} \quad & V_r^\pi(s_0)
\label{eq:obj}\\
\text{s.t.} \quad&
V_c^\pi(h)\le \theta(s_{t})
& \text{ for all } h = (s_0,a_0,\dots,s_{t}),\,
t \ge 0: \mathbb{P}_\pi(h) > 0
\label{eq:asw-const}
\end{align}
The formulation imposes a constraint on every subprocess starting from a reachable history $h$.

\section{Why Not Stationary Policies?}
\label{sec:why-not-stationary}

In standard CMDPs, \emph{stationary} policies---which map each state to an action distribution---suffice for optimality.
This is no longer the case for S-CMDPs.
The maximum value attained by stationary policies can be strictly lower than that attained by history-dependent policies. Moreover, even if we accept this loss of optimality and restrict attention to stationary policies, optimizing over this class is computationally intractable. Both limitations persist in the finite-horizon setting.

\begin{wrapfigure}[5]{r}{0.4\textwidth}
\vspace{-2mm}
\hspace{10mm}
\begin{tikzpicture}[->]
\node (s3) at (0,0) {$s_3$};
\node (s1) at (-1.2,0.5) {$s_1$};
\node (s2) at (-1.2,-0.5) {$s_2$};
\node (t)  at (1.5,0) {terminal};

% incoming arrows indicating initial states
\node (i1) at (-2,0.5) {~};
\node (i2) at (-2,-0.5) {~};

\draw[->] (i1) -- (s1);
\draw[->] (i2) -- (s2);

% transitions
\draw[->] (s1) -- (s3);
\draw[->] (s2) -- (s3);
\draw[->] (s3) -- (t);
\end{tikzpicture}
\end{wrapfigure}

\paragraph{Suboptimality of Stationary Policies}

Consider the S-CMDP shown on the right. The process starts in either $s_1$ or $s_2$ with equal probability.\footnote{For ease of exposition, we slightly deviate from the model in \Cref{sc:prelim} and allow a randomized initial state.}
There are two actions, $a$ and $b$.
All transitions are deterministic, as depicted in the figure, and are independent of the chosen action.
All rewards and costs are zero except for $r(s_3,b)=c(s_3,b)=1$.
The cost thresholds are $\theta(s_1)=0$ and $\theta(s_2)=\theta(s_3)=1$.

The only way to obtain positive reward in this instance is to choose action $b$ at $s_3$, which also incurs cost $1$. Under any stationary policy, however, choosing $b$ at $s_3$ with positive probability would violate the subprocess constraint at $s_1$, where the onward cost must not exceed $\theta(s_1)=0$. Hence, every feasible stationary policy must choose action $a$ at $s_3$ with probability $1$, yielding value zero.\footnote{In this example, even allowing the policy to be time-dependent does not improve the outcome.}

In contrast, history dependence allows the agent to condition its action at $s_3$ on the preceding state: choose $a$ if the trajectory starts from $s_1$, and choose $b$ if it starts from $s_2$. This satisfies all subprocess constraints and yields expected value $\gamma/2$, strictly outperforming all feasible stationary policies.

% ---------------------------------------
% - global tikzset
% ---------------------------------------
\colorlet{myblue}{blue!60!cyan}
    
\tikzset{
->, % makes the edges directed
% >={Stealth[scale=1.0]}, % makes the arrow heads bold
node distance=1.7cm, 
every state/.style={semithick, fill=white, minimum size=7mm},
every edge/.append style={semithick},
initial text= ,
}

\tikzstyle{edgeLabel}=[fill=white,minimum height=2mm,minimum width=4mm,inner sep=0.5mm]
\tikzstyle{terminalState}=[state, accepting, minimum size=5mm]

\paragraph{Intractability of Stationary Policies}

A further limitation of stationary policies is computational intractability. Optimizing over feasible stationary policies, even approximately, is NP-hard. This intractability persists even when the subprocess constraints are relaxed to permit small violations.
We formalize this claim with the following approximation notion and \Cref{thm:hardness}.

\begin{definition}[$(\epsilon,\delta)$-optimal policy]
\label{def:approx-opt}
A policy $\pi$ is $(\epsilon,\delta)$-optimal if the following conditions hold:
\begin{itemize}
\item[1)] {\bf $\delta$-feasibility:} $\pi$ violates every subprocess constraint in \Cref{eq:asw-const} by at most $\delta$. 
\item[2)] {\bf $\epsilon$-optimality:} $V_r^\pi(s_0) \ge V_r^{\pi'}(s_0) - \epsilon$ for every policy $\pi'$ in the class under consideration that satisfies \Cref{eq:asw-const} exactly.
\end{itemize}
Unless stated otherwise, the class contains all policies. In \Cref{thm:hardness}, both the computed policy and the policies $\pi'$ used for comparison are stationary.
\end{definition}

\begin{restatable}{theorem}{thmhardness}
\label{thm:hardness}
Fix any discount factor $\gamma \in (0,1)$. There exist constants $\epsilon,\delta>0$ such that computing a {\bf\em stationary} policy that is $(\epsilon,\delta)$-optimal among stationary policies is NP-hard.
\end{restatable}

For the remainder of the paper, we turn to history-dependent policies and show that they overcome both limitations of stationary policies.

\section{Enabling History Dependence: Two Birds with One Stone}
\label{sec:histoy-dependence}

While allowing history dependence removes the restriction behind the preceding suboptimality example, its computational tractability is far from obvious. Because the number of reachable histories can grow exponentially with the horizon, even writing down a history-dependent policy may be infeasible, let alone computing an optimal one.

We therefore adopt an implicit computational view and ask:
{\em given any history $h$, can we efficiently compute the action distribution of a near-optimal policy at $h$?}
An algorithm that does so allows us to execute the policy without explicitly storing all action distributions; effectively, it encodes the policy.

The value-set iteration (VSI) framework provides an answer to this question.
In this section, we first introduce a central concept underlying the VSI framework---the \emph{achievable value sets}---and explain how it enables us to handle history-dependent policies.

\subsection{Achievable Value Sets}

Our goal is to characterize which combinations of cumulative reward and cost can be attained by feasible policies. 
To this end, we view each state--action pair as producing a two-dimensional payoff whose components are reward and cost.
Formally, define a payoff function $\xx: S \times A \to \mathbb{R}^2$ such that
\begin{align*}
    \xx(s, a) := \left( r(s,a),\, c(s,a) \right)
\end{align*}
for every $(s,a) \in S \times A$.
Similarly, for any policy $\pi$ and history $h$, we write $\VV^\pi(h) = \left(V_r^\pi(h), V_c^\pi(h) \right)$ for the vector of onward discounted reward and cost.

\begin{definition}[Achievable value set]
The {\em achievable value set} of a state $s$, denoted $\calV^\star(s)$, is the set of all vectors $\vv \in \mathbb{R}^2$ that can be achieved by some {\em feasible} policy $\pi$ when starting from $s$, i.e.,
$\vv = \VV^\pi(s)$ for some $\pi$ satisfying \Cref{eq:asw-const}.
\end{definition}

We call functions like $\calV^\star$, that map each state to a subset of $\mathbb{R}^2$, {\em value-set functions}.

Using $\calV^\star$, we can express the optimal objective value in \Cref{eq:obj,eq:asw-const} as the maximum reward component over the achievable set at $s_0$:
\[
\max_{\vv = (v_1, v_2) \in \calV^\star(s_0)} v_1.
\]

If $\calV^\star$ were known, this maximization would give the optimal value immediately.
To further obtain a policy that achieves the optimal value, we next introduce a procedure that unrolls any target value $\vv \in \calV^\star(s)$ into a feasible policy $\pi$ with $\VV^\pi(s) = \vv$.
We will return to the computation of $\calV^\star$ and address the remaining algorithmic challenges in the following sections.

\subsection{Unrolling Policies Based on Achievable Sets}

Consider an arbitrary state $s$ and a target value vector $\vv = (v_1, v_2) \in \calV^\star(s)$.
By definition, some feasible policy $\pi$ achieves $\vv$, so a Bellman-style expansion gives
\begin{align}
\label{eq:const-bellman}
&\vv = 
\E_{a \sim \barpi} \Big( \xx(s, a) + \gamma \E_{s' \sim p(\cdot \given s, a)} \ww(a, s') \Big).
\end{align}
Here $\barpi = \pi(s) \in \Delta(A)$ is the action distribution at state $s$; and for each pair $(a,s')$, $\ww(a, s') = \VV^{\pi}(s, a, s')$ is the onward value, given history $(s, a, s')$.

Crucially, the fact that each onward value $\ww(a, s')$ is achieved in the subsequent process means that:
\begin{align}
\label{eq:const-onward}
\ww(a, s') \in \calV^\star(s') \;\; \text{for all } a, s': \barpi(a) \cdot p(s' \given s,a) > 0,
\end{align}
where only the $(s,a,s')$ triples with positive probability under $\barpi$ and the transition function $p$ are relevant, as the others do not contribute to \Cref{eq:const-bellman}.

\paragraph{Policy Execution via Recursive Value Expansion}

\Cref{eq:const-bellman,eq:const-onward} lead to the recursive value expansion procedure in \Cref{alg:achieve}.
The procedure expands a given target value $\vv \in \calV^\star(s)$ into an action distribution $\barpi$ and a collection $\ww$ of onward values based on \Cref{eq:const-bellman,eq:const-onward}. 
It samples an action $a$ from $\barpi$ and observes the next state $s'$.
The onward value $\ww(a,s')$ is then chosen as the new target, and the procedure recurses.
We fix a deterministic rule for selecting a valid decomposition whenever several are available.

\begin{wrapfigure}[14]{r}{0.45\textwidth}
\vspace{-14pt}

\begin{algorithm}[H]
\setstretch{1.2}
\small

\Fn{$\Expand(\vv,\, s)$}{
\smallskip

{\bf Return} $(\barpi, \ww)$ satisfying \Cref{eq:const-bellman,eq:const-onward}.

}

\caption{\Expand.}
\label{alg:expand}
\end{algorithm}

\vspace{4mm}

\begin{algorithm}[H]
\setstretch{1.2}

\Fn{$\Achieve(\vv,\, s)$}{

\smallskip

    1. $(\barpi, \ww) \leftarrow \Expand(\vv, s)$.
    
    2. Sample $a \sim \barpi$; observe $s' \sim p(\cdot \given s, a)$.

    3. \Achieve$(\vv', s')$, where $\vv' \!=\! \ww(a, s')$.
}

\caption{\Achieve.}
\label{alg:achieve}
\end{algorithm}

\end{wrapfigure}

Since $\vv \in \calV^\star(s)$, there always exists a valid decomposition into $\barpi$ and $\ww$. Moreover, \Cref{eq:const-onward} ensures that the new target value also lies within the corresponding achievable set, so a valid decomposition exists at the next recursive call as well. Thus, at every recursive call with current state $s$, the target value lies in $\calV^\star(s)$, so the procedure can continue indefinitely.

We can verify that the recursive procedure has expected discounted reward and cost exactly equal to $\vv$.
Effectively, the procedure implements a policy that achieves $\vv$ (i.e., $\VV^\pi(s)=\vv$).

Notably, executing this policy requires storing only the current target value. At each step, the target is expanded into 1) an action distribution used to sample the next action and 2) onward values used to update the target. Thus, the target serves as a fixed-dimensional summary of the history, independent of the horizon length.%
\footnote{The precision required to represent the exact target value may grow over time; the approximation approach in \Cref{sec:apprx-vsi} addresses this issue.}

\begin{wrapfigure}[8]{r}{0.45\textwidth}
\vspace{-20pt}

\begin{algorithm}[H]
\setstretch{1.2}
\small

\Fn{$\ActProbs(\vv,\, h = (s_0, a_0, \dots, s_t))$}{

\smallskip

1. $(\barpi, \ww) \leftarrow \Expand(\vv, s_0)$.

2. If $t=0$, then {\bf return} $\barpi$.

3. {\bf Return} $\ActProbs(\vv',\, h')$,
where $\vv' = \ww(a_0, s_1)$ and $h' = (s_1,a_1, \dots, s_t)$.\footnotemark

}

\caption{\ActProbs.}
\label{alg:act-probs}
\end{algorithm}
\end{wrapfigure}
\footnotetext{If $\barpi(a_0)\,p(s_1\given s_0,a_0)=0$ at any recursive step, then the input history is unreachable under the induced policy. In this case, $\ActProbs$ returns an arbitrary fixed distribution in $\Delta(A)$ without evaluating $\ww(a_0,s_1)$.}

\paragraph{Computing the Policy}

Returning to the algorithmic question posed at the beginning of this section, we can adapt the recursive expansion procedure to compute the action distribution $\pi(h)$ for any history $h$ starting at $s_0$ and any target $\vv \in \calV^\star(s_0)$, where $\pi$ is a policy that achieves $\vv$.
We present this algorithm as $\ActProbs$ in \Cref{alg:act-probs}.

Rather than sampling actions and states, $\ActProbs$ follows the given history $h$. On reachable histories, its recursive calls return the action distribution induced by $\Achieve(\vv,s_0)$ conditional on $h$.

Let $\pi$ be the policy computed by $\ActProbs(\vv, \cdot)$, i.e., $\pi(h) \coloneqq \ActProbs(\vv,h)$ for every history $h$ starting at $s_0$. It can be verified that $\pi$ is feasible and $\VV^\pi(s_0) = \vv$.
Hence, choosing $\vv^* \in \argmax_{\vv \in \calV^\star(s_0)} v_1$ as the target yields an optimal policy for the S-CMDP.

The following sections formalize these claims.
In summary, the algorithmic challenge of S-CMDPs now reduces to:
\begin{itemize}
\item[1)] computing $\calV^\star$, and
\item[2)] computing $\barpi$ and $\ww$ that satisfy \Cref{eq:const-bellman,eq:const-onward}.
\end{itemize}
% 1) computing $\calV^\star$, and 2) computing $\barpi$ and $\ww$ that satisfy \Cref{eq:const-bellman,eq:const-onward}.
We introduce the VSI algorithm next.

\section{Value-Set Iteration}
\label{sec:vsi}

At a high level, our approach mirrors classical value iteration for unconstrained MDPs: we define an operator whose \emph{fixed point} characterizes the achievable values, and we design an iterative process that converges to this fixed point. In our setting, however, the operator acts on sets of vectors rather than scalar values. More crucially, due to the subprocess constraints, its convergence requires a fundamentally different argument from that used for classical value iteration.

\subsection{$\calV^\star$ as a Fixed Point}

To begin, let us characterize $\calV^\star$ as a fixed point.
Recall from \Cref{eq:const-bellman,eq:const-onward} that every $\vv \in \calV^\star(s)$ admits a Bellman-style decomposition.
In our S-CMDP model, there is one additional necessary condition: at state $s$, the cost component of every achievable vector $\vv$ must not exceed the state-wise threshold:
\begin{align}
\label{eq:const-state-wise}
v_2 \le \theta(s).
\end{align}
This condition is not just necessary: together with \Cref{eq:const-bellman,eq:const-onward}, it is sufficient for the achievability of $\vv$, as the following lemma states.

\begin{restatable}{lemma}{lmmcharacterization}
\label{lmm:characterization}
For any $s \in S$ and $\vv = (v_1,v_2) \in \mathbb{R}^2$, we have $\vv \in \calV^\star(s)$ if and only if
$v_2 \le \theta(s)$ and there exist a distribution $\barpi \in \Delta(A)$ and a vector $\ww(a,s') \in \mathbb{R}^2$ for each $(a,s') \in A \times S$ such that \Cref{eq:const-bellman,eq:const-onward} hold.
\end{restatable}

\Cref{lmm:characterization} characterizes $\calV^\star$ as a fixed point of the following operator $\Phi$ on value-set functions. For any $s \in S$,
\begin{align}
\label{eq:Phi}
& \Phi(\calV)(s) := 
\left\{\;\; 
\vv \in \mathbb{R}^2 \;\;\middle|\;\; 
\begin{array}{l}
\vv = \bellman(s, \barpi, \ww),\; 
\\[1mm]
(\barpi, \ww) \in \onward(s, \calV), \;
\\[1mm] 
v_2 \le \theta(s)
\end{array}\;\;
\right\},
\end{align}
where we use the shorthand $\bellman(s, \barpi, \ww)$ to represent the right-hand side of \Cref{eq:const-bellman},
and 
\begin{align}
\label{eq:onward}
\onward(s, \calV) := 
\left\{ (\barpi, \ww) \,\middle| 
\begin{array}{l}
\ww(a, s') \in \calV(s')
\;\text{ if }\; 
\barpi(a) \cdot p(s' \given s,a) > 0
\end{array}
\right\}
\end{align}
consists of all $(\barpi, \ww)$ pairs satisfying the onward achievability condition for every $(a,s') \in A \times S$, with respect to $\calV$ (i.e., \Cref{eq:const-onward} with $\calV^\star$ replaced by $\calV$).

Intuitively, $\Phi(\calV)(s)$ consists of values obtained by combining an action distribution with onward values from $\calV$, subject to the cost threshold at $s$.
\Cref{lmm:characterization} immediately yields the following corollary.

\begin{corollary}
\label{crl:fixed-point}
$\calV^\star$ is a fixed point of $\Phi$, i.e.,
$\calV^\star = \Phi(\calV^\star)$.
\end{corollary}

\subsection{Convergence of Value-Set Iteration to $\calV^\star$}

This suggests computing a fixed point of $\Phi$ to obtain $\calV^\star$.
For discounted unconstrained MDPs, the Bellman optimality operator admits a unique fixed point, which is exactly the optimal value function.
The existence and uniqueness of this fixed point follow from the contraction mapping theorem (Banach's fixed-point theorem), thanks to the Bellman operator being a {\em contraction}.\footnote{A mapping $T$ on a metric space $(\mathcal X,d)$ is a contraction if there exists $\beta\in[0,1)$ such that $d(T(x),T(y))\le \beta\, d(x,y)$ for all $x,y \in \mathcal X$.
}
Thus, through repeated application of the Bellman operator, classical value iteration converges to the fixed point at a geometric rate.

\begin{wrapfigure}[7]{r}{0.45\textwidth}
\vspace{-12pt}

\begin{algorithm}[H]
\setstretch{1.2}
\small

\Fn{$\VSI(\Phi, \calV)$}{

\smallskip

1. Repeat $\calV \leftarrow \Phi(\calV)$ until $\calV = \Phi(\calV)$.

2. {\bf Return} $\calV$.

}

\caption{$\VSI$ (value-set iteration).}
\label{alg:vsi}
\end{algorithm}

\end{wrapfigure}

Motivated by the above, it is tempting to design a similar iteration scheme by repeatedly applying $\Phi$ to a value-set function (\Cref{alg:vsi}), in the hope that it converges to $\calV^\star$.

This approach, however, requires more care.
Unlike the classical Bellman operator, $\Phi$ need not be a contraction, so the contraction mapping theorem does not apply in general.
Its fixed point may not be unique;\footnote{For example, assigning the empty set to every state gives a trivial fixed point, distinct from $\calV^\star$ whenever a feasible policy exists from at least one state.}
moreover, repeated application of $\Phi$ from an arbitrary initial value-set function may fail to converge or may converge to an undesired fixed point.

As it turns out, the following {\em monotonicity} property of $\Phi$ becomes key to our convergence argument.

\begin{restatable}[Monotonicity]{lemma}{lmmmonotonicity}
\label{lmm:monotonicity}
If $\calV \subseteq \calV'$, then $\Phi(\calV) \subseteq \Phi(\calV')$.%
\hideinrestate{\footnote{For simplicity, we use $\subseteq$ to denote the pointwise inclusion: e.g., $\calV \subseteq \calV'$ when $\calV(s) \subseteq \calV'(s)$ for all $s \in S$.}}
\end{restatable}

\paragraph{Convergence by Monotonicity}

The monotonicity property motivates a fixed-point argument inspired by Tarski's fixed-point theorem \cite{tarski1955lattice}. The theorem states that the fixed points of a monotone self-map on a complete lattice form a complete lattice and, in particular, include a \emph{greatest} fixed point.
In our setting, $\calV^\star$ is precisely the greatest bounded fixed point of $\Phi$: it contains every bounded fixed point of $\Phi$ (as we establish directly in \Cref{lmm:greatest}).

More importantly, monotonicity suggests a way to find this fixed point. If we initialize value-set iteration (\Cref{alg:vsi}) from an upper bound $\calV$ of $\calV^\star$ that $\Phi$ maps into itself, i.e., $\Phi(\calV) \subseteq \calV$, we obtain a decreasing sequence of value-set functions.
Indeed, since all rewards and costs lie in $[-1,1]$,
the $L_\infty$ ball
\[
\textstyle
\calB := 
\left\{ 
\vv \in \mathbb{R}^2 \,\middle|\, 
\|\vv\|_\infty \le \frac{1}{1 - \gamma}
\right\}
\]
provides such an upper bound, as the following lemma states.

\begin{restatable}{lemma}{lmmballproperties}
\label{lmm:ball-properties}
$\calV^\star \subseteq \calB$ and $\Phi(\calB) \subseteq \calB$.%
\hideinrestate{\footnote{When we apply $\Phi$ to $\calB$, we interpret $\calB$ as the constant value-set function mapping every state to the ball $\calB$.}}
\end{restatable}

We show that the iterates of $\VSI(\Phi, \calB)$ converge to $\calV^\star$.
Let $(\calV^k)_{k \ge 0}$ be the sequence it generates:
\begin{itemize}
\item $\calV^0 = \calB$, and
\item $\calV^{k+1} = \Phi(\calV^k)$ for all $k \ge 0$.
\end{itemize}
% \[
% \calV^0 = \calB,
% \quad\text{and}\quad
% \calV^{k+1} = \Phi(\calV^k) \text{ for all } k \ge 0.
% \]
By \Cref{lmm:ball-properties}, $\calV^\star \subseteq \calB$, so monotonicity of $\Phi$ gives $\Phi(\calV^\star)\subseteq \Phi(\calB)$.
Since $\calV^\star = \Phi(\calV^\star)$ and $\Phi(\calB) \subseteq \calB$, we obtain
\[
\calV^\star \subseteq \Phi(\calB) \subseteq \calB.
\]
Recall that $\calV^0=\calB$ and $\calV^1=\Phi(\calV^0)$, so this becomes
\[
\calV^\star \subseteq \calV^1 \subseteq \calV^0.
\]
By monotonicity, repeatedly applying $\Phi$ to these inclusions propagates the nesting to all $k$, yielding the following result.

\begin{restatable}{lemma}{lmmnested}
\label{lmm:nested}
$\calV^\star \subseteq \calV^{k+1} \subseteq \calV^k$ for all $k \ge 0$.
\end{restatable}

Consequently, $(\calV^k)_{k \ge 0}$ is a decreasing sequence bounded below by $\calV^\star$. Each $\calV^k(s)$ is compact, and a compactness argument shows that the intersection of these iterates is a fixed point of $\Phi$ (see the proof of \Cref{thm:convergence}).
This fixed point must equal $\calV^\star$, which bounds it not only from below but also, by the following lemma, from above.

\begin{restatable}{lemma}{lmmgreatest}
\label{lmm:greatest}
If $\calV$ is bounded and $\calV = \Phi(\calV)$, then $\calV \subseteq \calV^\star$.
\end{restatable}

The proof of \Cref{lmm:greatest} follows the recursive construction in \Cref{sec:histoy-dependence}, where we showed that every $\vv \in \calV^\star(s)$ can be achieved by the $\Achieve$ procedure (\Cref{alg:achieve}). Here, we show more generally that for any bounded fixed point $\calV$ of $\Phi$, every $\vv \in \calV(s)$ can be achieved by the same procedure, with $\calV$ in place of $\calV^\star$ in the onward value constraint in \Cref{eq:const-onward}.
This modified procedure is well defined because $\calV=\Phi(\calV)$ guarantees a valid decomposition at every recursive call. Consequently, every $\vv \in \calV(s)$ is realized as the expected discounted return of a feasible policy and is therefore achievable from $s$.

These lemmas lead to the following convergence theorem.

\begin{restatable}{theorem}{thmconvergence}
\label{thm:convergence}
$\calV^k \to \calV^\star$ as $k \to \infty$.% 
\hideinrestate{\footnote{More precisely, we have $\calV^\infty = \calV^\star$, where $\calV^\infty(s) := \bigcap_{k \ge 0} \calV^k(s)$ for every $s \in S$.}}
\end{restatable}

\section{Approximate VSI for Finite-Time Convergence}
\label{sec:apprx-vsi}

We have established the convergence of value-set iteration to $\calV^\star$, but this does not automatically yield a finite-time algorithm, let alone an efficient one. To obtain a practical algorithm, two issues must be addressed:
1) ensuring termination in finitely---preferably polynomially---many iterations; and
2) representing value-set functions in a way that allows applications of $\Phi$ to be computed efficiently.

As in practical implementations of value iteration, we adopt an approximation approach. Rather than computing $\calV^\star$ exactly, we iterate an approximate operator that reaches a fixed point in finitely many steps. As we show in this section, this approach is sufficient for our purpose of computing an $(\epsilon,\delta)$-optimal policy (\Cref{def:approx-opt}).

\subsection{Approximate Operator and Convergence}

\begin{wrapfigure}[9]{r}{0.40\textwidth}
\vspace{-17pt}
\hspace{3mm}
\begin{tikzpicture}[scale=0.75,transform shape]
\tikzstyle{mydot} = [circle,fill,inner sep=0.9pt]

\draw[myblue!50, thick, fill=myblue!3] plot[smooth cycle, tension=0] coordinates{
(-2.5, -0.5) 
(-2.5,  0.0)
(-2.0,  1.0)
(-1.5,  1.5)
(-0.5,  2.0)
( 1.5,  2.0)
( 2.0,  1.5)
( 2.0,  0.0)
( 0.5, -1.5)
(-1.5, -1.5)
};  

\draw[black,thick, rotate=30] (0,0.25) ellipse (2cm and 1.5cm);

\draw[black,thick, dotted, rotate=30] (0,0.25) ellipse (2.6cm and 1.95cm);

\foreach \x in {0,...,12}
\foreach \y in {0,...,9}
  \fill[black!50] (\x*0.5-3,\y*0.5-2) circle (0.6pt);

% -----

\node[black] at (0, 0.2) {\large $\Phi(\calV)(s)$};
\node[myblue!90, fill=white] at (-2.3, 2) {\large $\xPhi(\calV)(s)$};
\node[black, fill=white] at ( 3, -1.5) {\large $\neigh_\xi(\Phi(\calV)(s))$};

\draw[myblue!90,-,semithick] (-2.2, 1.6) -- (-1.7, 1.0);
\draw[black,-,semithick] (1.8, -0.5) -- (2.3, -1.1);

\end{tikzpicture}
\vspace{-5.5mm}
\caption{Illustration of $\xPhi$.}
\label{fig:xPhi}
\end{wrapfigure}

Starting from $\calB$, repeated applications of $\Phi$ produce convex polytopes at each state (as the LP formulation in \Cref{sec:computing-eps-delta-opt} makes explicit). These convex polytopes can in principle be represented by linear inequalities, but the issue is that the number of inequalities can grow exponentially under repeated applications of $\Phi$.
To control the representation complexity, we discretize the two-dimensional value space using a grid and approximate $\Phi$ based on this discretization.

\paragraph{Approximate Operator}
Choose a positive grid spacing $\xi \le (1-\gamma) \cdot \min \{\epsilon,\, \delta/2\}$.
Let $G_\xi \coloneqq \{\xi \xx \given \xx \in \mathbb{Z}^2 \}$ be the set of grid points. For any $X \subseteq \mathbb{R}^2$, define
\[
    \gridhull_\xi(X)
    \coloneqq \conv\!\left(X \cap G_\xi\right),
\]
where $\conv(\cdot)$ denotes the convex hull. 
Define the approximate operator as
\begin{align}
\label{eq:xphi}
    \xPhi(\calV)(s)
    :=
    \gridhull_\xi\!\left(
        \neigh_\xi\!\left(\Phi(\calV)(s)\right)
    \right),
\end{align}
where $\neigh_\xi(X)$ denotes the $\xi$-neighborhood of $X$,
consisting of all points within $L_\infty$ distance $\xi$ of some point in $X$.

\Cref{fig:xPhi} illustrates this construction.
Whenever $\Phi(\calV)(s)$ is convex, its $\xi$-neighborhood is also convex, and we can verify that the following inclusions hold:
\[
    \Phi(\calV)(s)
    \subseteq
    \xPhi(\calV)(s)
    \subseteq
    \neigh_\xi\!\left(\Phi(\calV)(s)\right).
\]
Specifically, the left-hand inclusion holds even without convexity of $\Phi(\calV)(s)$: every point of $\Phi(\calV)(s)$ lies in a grid cell whose vertices are within $L_\infty$ distance $\xi$ of that point; the cell is contained entirely in $\neigh_\xi\!\left(\Phi(\calV)(s)\right)$ as well as its grid hull. The right-hand inclusion follows because the convex neighborhood contains the convex hull of its grid points.

\paragraph{Finite-Time Convergence}

The operator $\xPhi$ remains monotone. To initialize VSI, let $\xcalB_\xi$ denote the smallest grid-aligned $L_\infty$ ball that {\em strictly} contains $\calB$, i.e.,
\[
\textstyle
    \xcalB_\xi
    \coloneqq
    \left\{
        \vv \in \mathbb{R}^2
        \,\middle|\,
        \|\vv\|_\infty
        \le
        \left( 
            \left\lfloor 
                \frac{1}{1-\gamma} / \xi 
            \right\rfloor 
            + 1
        \right) \cdot \xi
    \right\}.
\]
We run $\VSI(\xPhi,\xcalV^0)$ (\Cref{alg:vsi}) with
$\xcalV^0 = \xcalB_\xi$.
This yields a sequence $(\xcalV^k)_{k \ge 0}$ such that:
\begin{itemize}
    \item $\xcalV^0 = \xcalB_\xi$, and
    \item $\xcalV^{k+1} = \xPhi(\xcalV^k)$ for all $k \ge 0$.
\end{itemize}
The strict outer rounding of $\xcalV^0$ ensures $\xPhi(\xcalV^0) \subseteq \xcalV^0$.
Together with $\calV^\star \subseteq \calB \subseteq \xcalV^0$ and the monotonicity of $\xPhi$, this makes
$(\xcalV^k)_k$ a decreasing sequence of nested value sets. We extend \Cref{lmm:nested} to this sequence as follows.

\begin{restatable}{lemma}{lmmnestedx}
\label{lmm:nested-x}
$\calV^\star \subseteq \xcalV^{k+1} \subseteq \xcalV^k$ for all $k \ge 0$.
\end{restatable}

Given \Cref{lmm:nested-x}, the discretization then ensures finite-time convergence.
Specifically, by construction, each $\xcalV^k(s)$ is fully characterized by the grid points it contains.
Since the sequence is decreasing, whenever $\xcalV^{k+1} \neq \xcalV^k$, at least one grid point must have been removed from the latter.
Let $N_\xi = |G_\xi \cap \xcalB_\xi|$ be the number of grid points inside $\xcalB_\xi$. Across the $|S|$ value sets, at most $|S|N_\xi$ grid points can be removed, so $\VSI(\xPhi,\xcalV^0)$ reaches a fixed point within $O(|S|N_\xi)$ iterations. When $\gamma$ and $\xi$ are fixed, $N_\xi$ is a constant, so the iteration bound is $O(|S|)$.

\subsection{$(\epsilon,\delta)$-Optimality at Convergence}

Let $\xcalV^\star$ be the fixed point obtained via $\texttt{VSI}(\xPhi,\, \xcalV^0)$, i.e., 
\[
\exists\, k \ge 0: \;
\xcalV^\star = \xcalV^k = \xcalV^{k+1}.
\]
By \Cref{lmm:nested-x}, $\calV^\star \subseteq \xcalV^\star$, so the approximation retains all achievable values, including the optimal one.

Although $\xcalV^\star$ may contain values outside $\calV^\star$, \Cref{lmm:eps-delta-achievable} shows that every value it contains is achievable {\em approximately}, in the sense defined below.
Choose $\vv^* \in \argmax_{\vv \in \xcalV^\star(s_0)} v_1$. The lemma gives a $\delta$-feasible policy whose return is within $\epsilon$ of $\vv^*$. Since $v_1^*$ is at least the reward of every exactly feasible policy, this policy is $(\epsilon,\delta)$-optimal (\Cref{def:approx-opt}).

\begin{definition}[$(\epsilon,\delta)$-achievability]
$\vv \in \mathbb{R}^2$ is {\em $(\epsilon,\delta)$-achievable} from state $s$ if there exists a $\delta$-feasible policy $\pi$, with $\|\VV^\pi(s) - \vv\|_\infty \le \epsilon$.
\end{definition}

\begin{restatable}{lemma}{lmmepsdeltaachievable}
\label{lmm:eps-delta-achievable}
If $\calV$ is bounded and $\calV = \xPhi(\calV)$, then for all $s \in S$ every $\vv \in \calV(s)$ is $(\epsilon,\delta)$-achievable from $s$.
\end{restatable}

The proof of \Cref{lmm:eps-delta-achievable} follows the same recursive construction as that of \Cref{lmm:greatest}, allowing a bounded error at each expansion. The corresponding extension of $\Achieve$ (\Cref{alg:achieve}), denoted $\Achieve_\xi$, replaces $\Expand$ with the relaxed subroutine $\Expand_\xi$ shown in \Cref{alg:expand-x}.

\begin{wrapfigure}[7]{r}{0.45\textwidth}
\vspace{-15pt}

\begin{algorithm}[H]
\setstretch{1.2}
\small

\Fn{$\Expand_\xi(\vv,\, s)$}
{
\smallskip

{\bf Return}   
$(\barpi, \ww) \in \onward(s, \xcalV^\star)$ such that\\ ~$\bellman(s, \barpi, \ww) \in \neigh_\xi(\vv)$.

}

\caption{$\Expand_\xi$.}
\label{alg:expand-x}
\end{algorithm}

\end{wrapfigure}

\paragraph{Relaxed Value Expansion}
Compared with the exact implementation, $\Expand_\xi$ only requires $(\barpi, \ww)$ to induce a value close to $\vv$. This relaxation is essential because $\xcalV^\star$ is a fixed point of $\xPhi$, but need not be a fixed point of the original operator $\Phi$. In general, it can contain values outside $\Phi(\xcalV^\star)$.

Indeed, insisting on an exact Bellman expansion of such values may require onward vectors outside $\xcalV^\star$. Repeating this at subsequent steps can cause the targets to drift outward and become unbounded, preventing them from being realized as discounted returns.

The procedure $\Expand_\xi$ prevents this drift by keeping the onward vectors within $\xcalV^\star$ while allowing a Bellman error of at most $\xi$ at each step. Such an expansion exists because $\xcalV^\star \subseteq \neigh_\xi(\Phi(\xcalV^\star))$.
Due to geometric discounting, the total accumulated offset is bounded by $\xi/(1-\gamma) \le \epsilon$.

Replacing $\Expand$ with $\Expand_\xi$ in \Cref{alg:act-probs} yields a corresponding extension of $\ActProbs$. The extension, call it $\ActProbs_\xi$, computes the action distributions prescribed by a $\delta$-feasible policy that achieves the target value within error $\epsilon$. We omit the pseudocode for both $\ActProbs_\xi$ and $\Achieve_\xi$.

\subsection{Computation and Tractability}
\label{sec:computing-eps-delta-opt}

It remains to make these operations computationally efficient. As outlined at the end of \Cref{sec:histoy-dependence}, we need to:
\begin{itemize}
\item[1)] 
compute applications of $\xPhi$, so that we can obtain its fixed point using $\VSI$; and

\item[2)]
compute $(\barpi, \ww)$ in $\Expand_\xi$, so that ultimately we can execute an $(\epsilon,\delta)$-optimal policy using $\Achieve_\xi$, or compute its action distributions using $\ActProbs_\xi$.
\end{itemize}
We establish tractability by reducing both tasks to linear programming (LP).

\paragraph{Computing $\xPhi(\calV)(s)$}
From the construction in \Cref{eq:xphi}, computing each $\xPhi(\calV)(s)$ reduces to determining, for each grid point $\mathbf{g}$, whether $\mathbf{g} \in \neigh_\xi(\Phi(\calV)(s))$. We can restrict attention to grid points in $\xcalB_\xi$ because by \Cref{lmm:nested-x} all relevant value sets are contained in $\xcalV^0 = \xcalB_\xi$. 
Thus, for fixed $\gamma$ and $\xi$, there is only a fixed number of grid points to check per state.

To efficiently decide whether $\mathbf{g} \in \neigh_\xi(\Phi(\calV)(s))$, we show that $\neigh_\xi(\Phi(\calV)(s))$ can be characterized by linear constraints, so the decision problem reduces to an LP.
Indeed, $\vv \in \neigh_\xi(\Phi(\calV)(s))$ means that there exists some $\tilde{\vv} \in \Phi(\calV)(s)$ such that
\begin{align}
    \|\vv-\tilde{\vv}\|_\infty
    &\le \xi.
    \label{eq:computing-dist}
\end{align}
According to the definition of $\Phi$ in \Cref{eq:Phi}, the membership $\tilde{\vv} \in \Phi(\calV)(s)$ is in turn equivalent to the existence of $\barpi \in \Delta(A)$ and $\ww$ satisfying the following constraints:
\begin{align}
    &\tilde{\vv}
    =
    \sum_{a \in A} \barpi(a)
    \left(
        \xx(s,a)
        +
        \gamma
        \sum_{s' \in S}
        p(s' \given s,a)\,\ww(a,s')
    \right),
    \label{eq:computing-bellman}
    \\[1mm]
    &\tilde v_2
    \le \theta(s),
    \label{eq:computing-threshold}
    \\[1mm]
    &\ww(a,s')
    \in \calV(s')
    \qquad
    \text{for all } a,s':
    \barpi(a)\cdot p(s' \given s,a)>0.
    \label{eq:computing-onward}
\end{align}
Hence,
$\vv \in \neigh_\xi(\Phi(\calV)(s))$ if and only if there exist
$\tilde{\vv}$, $\barpi \in \Delta(A)$, and $\ww$ satisfying
\Cref{eq:computing-dist,eq:computing-bellman,eq:computing-threshold,eq:computing-onward}.

These constraints are nonlinear in $\barpi$ and $\ww$: \Cref{eq:computing-bellman} contains products $\barpi(a)\cdot\ww(a,s')$, while the pairs constrained by \Cref{eq:computing-onward} depend on $\barpi(a)$.
To convert them into standard linear constraints, we first rewrite
\Cref{eq:computing-onward}, for pairs with $p(s'\given s,a)>0$ and $\calV(s')\neq\emptyset$, as
\[
    \barpi(a)\cdot\ww(a,s')
    \;\in\;
    \barpi(a)\cdot\calV(s'),
\]
where 
$\barpi(a)\cdot\calV(s')
\coloneqq \{\barpi(a)\cdot\vv \mid \vv\in\calV(s')\}$
is the polytope $\calV(s')$ scaled by $\barpi(a)$. If $p(s'\given s,a)>0$ but $\calV(s')=\emptyset$, feasibility instead requires $\barpi(a)=0$.
For nonempty successor sets, each product $\barpi(a)\cdot\ww(a,s')$ is a convex combination of the scaled vertices. We introduce convex weights to eliminate $\ww(a,s')$ and introduce auxiliary variables to linearize the products between $\barpi(a)$ and these weights.

More formally, we present the following lemma, establishing the tractability of $\xPhi(\calV)(s)$.

\begin{restatable}{lemma}{lmmlinearize}
\label{lmm:linearize}
Suppose each $\calV(s')$ is given as the convex hull of finitely many grid points. For any $s$ and $\vv$, the feasibility of \Cref{eq:computing-dist,eq:computing-bellman,eq:computing-threshold,eq:computing-onward} can be decided by solving an LP.
\end{restatable}

\begin{algorithm}[t]

% \setstretch{1.05}

\begin{itemize}[leftmargin=1em, itemsep=1pt, labelsep=0.3em]
\item[1.]
Run $\VSI(\xPhi, \xcalV^0)$ to obtain a fixed point $\xcalV^\star$, where $\xcalV^0 = \xcalB_\xi$.
If $\xcalV^\star(s_0)=\emptyset$, report that no exactly feasible policy exists and stop.

\item[2.]
Choose a target $\vv^* \in \argmax_{\vv \in \xcalV^\star(s_0)} v_1$.

\item[3.]
{\bf To execute a policy:} Run $\Achieve_\xi(\vv^*,\, s_0)$ to achieve $\vv^*$ within $\epsilon$ error while violating each subprocess constraint by at most $\delta$.

{\bf To compute a policy:} Run $\ActProbs_\xi(\vv^*,\, h)$ to compute $\pi(h)$ of an $(\epsilon, \delta)$-optimal policy $\pi$.
\end{itemize}
\vspace{-3mm}
\caption{VSI framework for solving S-CMDPs.}
\label{alg:summary}
\end{algorithm}

\paragraph{Computing $\Expand_\xi(\vv,s)$}
For any $\vv \in \xcalV^\star(s)$, a valid output of $\Expand_\xi(\vv,s)$ can be obtained by solving \Cref{eq:computing-dist,eq:computing-bellman,eq:computing-onward} with $\calV=\xcalV^\star$ and the given $\vv$.
The same linearization as above applies directly, so the tractability of $\Expand_\xi(\vv,s)$ follows.

\medskip

Consequently, we establish \Cref{thm:poly-time} as a main algorithmic result.
An overview of the VSI framework is given in \Cref{alg:summary}.

\begin{restatable}{theorem}{thmpolytime}
\label{thm:poly-time}
For fixed $\gamma\in(0,1)$, choosing $\xi=(1-\gamma)\min\{\epsilon,\delta/2\}$ allows us to compute a fixed point $\xcalV^\star \supseteq \calV^\star$ of $\xPhi$ in time polynomial in the input size, $1/\epsilon$, and $1/\delta$. If $\xcalV^\star(s_0)\neq\emptyset$, there exists an $(\epsilon,\delta)$-optimal policy $\pi$ whose action distribution $\pi(h)$ can be computed in time polynomial in the same quantities and $|h|$ for any given history $h$.
\end{restatable}

\section{Implementation}
\label{sec:implementation}

To implement the VSI framework, we introduce one main practical adaptation: a mixed-integer linear programming (MILP)-based procedure for computing $\xPhi(\calV)$.

In the theoretical algorithm, each value set $\xPhi(\calV)(s)$ is obtained by exhaustively checking every grid point in the bounded value region. Although this approach suffices to establish tractability, it can be inefficient in practice because many grid points---particularly those in the interior of the polytope---are unnecessary for its vertex representation. Our implementation instead employs an iterative MILP-based separation procedure that searches directly for grid points that expand the current convex hull.

\begin{algorithm}[t]
\DontPrintSemicolon
\caption{An MILP-based practical algorithm for computing vertices of $\xPhi(\calV)(s)$.}
\label{alg:vertices-milp}

\small

\setstretch{1.1}

\KwIn{State $s\in S$; current value-set function $\calV$ in vertex representation}
\KwOut{Vertex representation of $\xPhi(\calV)(s)$}

\BlankLine

Solve an LP to find an arbitrary $\vv \in \Phi(\calV)(s)$;

\BlankLine

\If{LP infeasible}{
    \Return $\emptyset$;
}

\BlankLine
Initialize $\mathcal{K}$ as the set of vertices of the grid cell containing $\vv$;

\BlankLine
\Repeat{no new grid point is found}{

    \smallskip
    Compute the convex hull of $\mathcal{K}$;

    \smallskip
    
    \For{each facet of the hull}{

        \smallskip
        Let $\nn$ be the outward normal vector of the facet;

        \smallskip
        
        Solve the MILP:\
        $\max_{\mathbf{g} \in G_\xi} \, \nn \cdot \mathbf{g}$,\ 
        subject to
        $\mathbf{g} \in \neigh_\xi \big(\Phi(\calV)(s) \big)$;
        \tcp*[r]{$G_\xi=\{ \xi\xx \mid \xx \in \mathbb{Z}^2 \}$~}

        \smallskip
        
        \If{MILP returns a grid point $\mathbf{g}$ outside the current hull}{
            \smallskip
            $\mathcal{K} \leftarrow \mathcal{K} \cup \{\mathbf{g} \}$;

            \textbf{break};
        }
    }
}

\BlankLine
\Return $\mathcal{K}$.
\end{algorithm}

% Change this width to resize the panels in all three implementation figures.
\newcommand{\implementationpanelwidth}{0.235\linewidth}
% Crops omit axis titles while retaining tick labels and state titles.

\begin{figure}[t]
    \centering
    \begin{tabular}{@{}c@{\hspace{0.01\linewidth}}c@{\hspace{0.01\linewidth}}c@{\hspace{0.01\linewidth}}c@{}}
    \includegraphics[width=\implementationpanelwidth,trim={40pt 42pt 877pt 30pt},clip]{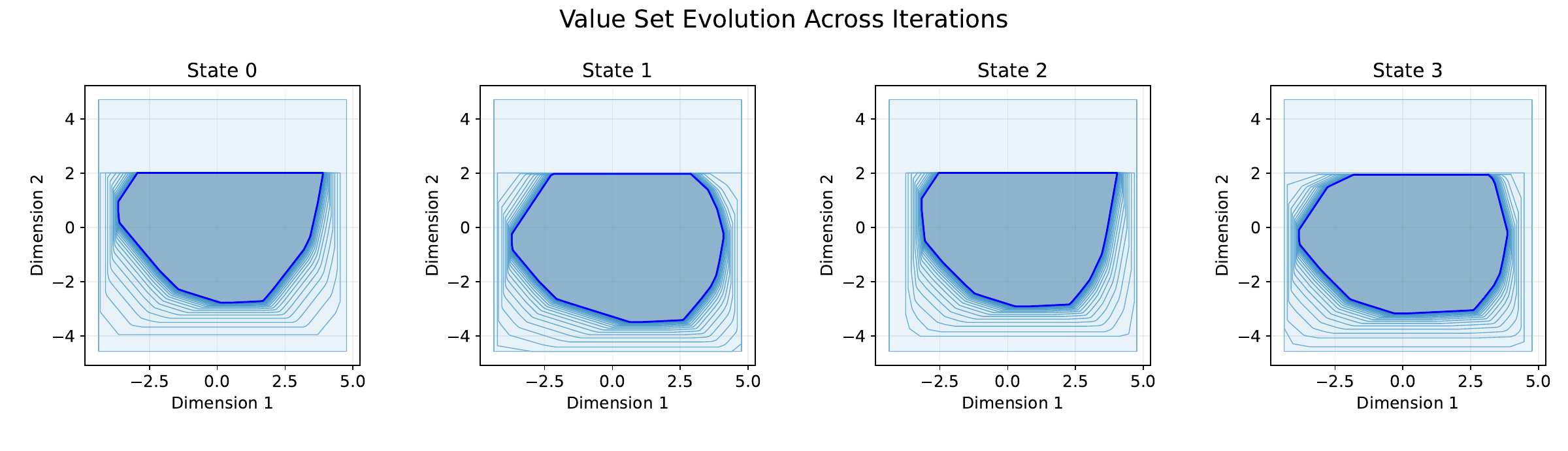} &
    \includegraphics[width=\implementationpanelwidth,trim={330pt 42pt 592pt 30pt},clip]{experiments/iteration_snapshots.pdf} &
    \includegraphics[width=\implementationpanelwidth,trim={620pt 42pt 301pt 30pt},clip]{experiments/iteration_snapshots.pdf} &
    \includegraphics[width=\implementationpanelwidth,trim={911pt 42pt 11pt 30pt},clip]{experiments/iteration_snapshots.pdf}
    \end{tabular}
    \caption{\small
    Value sets from an example run. The horizontal and vertical axes show expected discounted reward and cost, respectively. In each plot, the dark inner polytope represents the final value set. The lighter outer polytopes represent intermediate value sets generated during the iterations. For this instance, VSI converges after 37 iterations.
    }
    \label{fig:snapshots}

    % \vspace{3mm}

    \begin{tabular}{@{}c@{\hspace{0.01\linewidth}}c@{\hspace{0.01\linewidth}}c@{\hspace{0.01\linewidth}}c@{}}
    \includegraphics[width=\implementationpanelwidth,trim={27pt 69pt 872pt 30pt},clip]{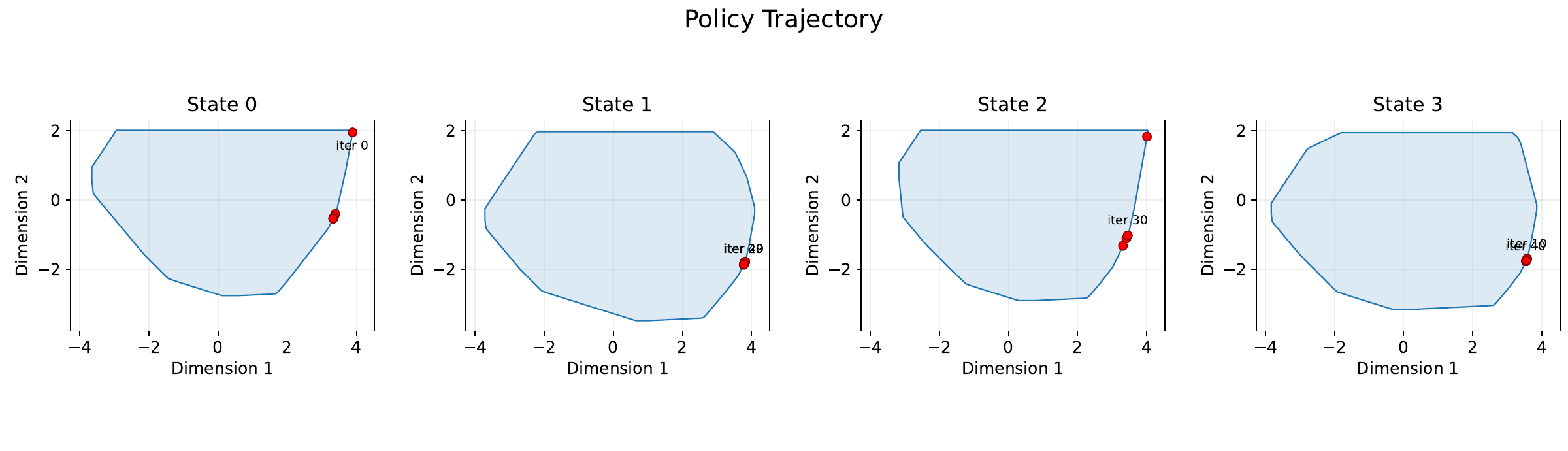} &
    \includegraphics[width=\implementationpanelwidth,trim={317pt 69pt 581pt 30pt},clip]{experiments/policy_trajectory.pdf} &
    \includegraphics[width=\implementationpanelwidth,trim={608pt 69pt 291pt 30pt},clip]{experiments/policy_trajectory.pdf} &
    \includegraphics[width=\implementationpanelwidth,trim={898pt 69pt 0pt 30pt},clip]{experiments/policy_trajectory.pdf}
    \end{tabular}
    \caption{\small
    Trajectory of target values generated during the execution of an $(\epsilon,\delta)$-optimal policy for the same instance as in \Cref{fig:snapshots}. The horizontal and vertical axes show expected discounted reward and cost, respectively. The red points mark the target values.
    }
    \label{fig:trajectory}
\end{figure}

\subsection{Vertex Computation via MILP Separation}

By definition,
\[
    \xPhi(\calV)(s)
    =
    \gridhull_\xi\!\left(
        \neigh_\xi\!\left(\Phi(\calV)(s)\right)
    \right),
\]
so computing $\xPhi(\calV)(s)$ amounts to computing the convex hull of the grid points contained in $\neigh_\xi\!\left(\Phi(\calV)(s)\right)$. 
Under a vertex representation,%
\footnote{We use ``vertex representation'' slightly loosely to allow redundant points that are not necessarily extreme points. An irredundant representation can be obtained by extracting the vertices of $\conv(\mathcal{K})$ at the end of \Cref{alg:vertices-milp}, but this is not necessary for the algorithm.}
it suffices to identify a set of grid points whose convex hull equals $\xPhi(\calV)(s)$.

To avoid exhaustively enumerating all candidate grid points, our implementation builds on the LP formulation from \Cref{sec:computing-eps-delta-opt}.
For vertex discovery, we use the same linear constraints that characterize membership in
$\neigh_\xi(\Phi(\calV)(s))$, but additionally restrict the candidate value to $G_\xi$ using integer grid indices and optimize a linear
objective over the resulting feasible grid points. This yields an MILP that can search directly for grid points outside the current convex hull.
The search procedure, presented in
\Cref{alg:vertices-milp}, works as follows.
\begin{itemize}
\item[1.]
\textbf{Initialization.}
First, find an arbitrary point
$\vv \in \Phi(\calV)(s)$ by solving the LP from \Cref{sec:computing-eps-delta-opt} without the $\xi$-neighborhood relaxation. (Effectively, solve \Cref{eq:computing-bellman,eq:computing-threshold,eq:computing-onward} for $\tilde{\vv}$.)
If the LP is infeasible, then $\Phi(\calV)(s)=\emptyset$ and hence $\xPhi(\calV)(s)=\emptyset$, which we return. Otherwise, we use a grid cell containing $\vv$ as the initial polytope.

For every vertex $\mathbf{g}$ of this cell,
$\|\mathbf{g} - \vv\|_\infty \le \xi$. Since
$\vv \in \Phi(\calV)(s)$, we therefore have
\[
    \mathbf{g}
    \in
    \neigh_\xi\!\left(\Phi(\calV)(s)\right)
    \cap G_\xi.
\]
Hence, by the definition of $\gridhull_\xi$, the entire initial cell is contained in $\xPhi(\calV)(s)$. Using the cell ensures that the initial polytope is full-dimensional, so that the subsequent facet-enumeration procedure is well-defined.

\item[2.]
\textbf{Vertex discovery via facet separation.}
The algorithm then enumerates the facets of the current polytope. For each facet, it solves an MILP to optimality in the corresponding outward normal direction over all feasible grid points. The MILP uses the constraints from \Cref{eq:computing-dist,eq:computing-bellman,eq:computing-threshold,eq:computing-onward}, together with integer variables that restrict the candidate value to $G_\xi$.
\begin{itemize}
\item
If the maximizing grid point lies strictly beyond the current facet, that point is added to the maintained point set, the convex hull and its facets are recomputed, and the procedure repeats.

\item
If optimization over every facet direction produces no point outside the current polytope, the current polytope is returned as $\xPhi(\calV)(s)$.
\end{itemize}
\end{itemize}

The algorithm terminates because $\neigh_\xi(\Phi(\calV)(s))$ is bounded and therefore contains only finitely many grid points, while every non-terminal iteration adds a feasible grid point lying outside the current convex hull. 
The returned polytope is correct because a point outside a full-dimensional polytope must violate at least one of its facet-defining inequalities. At termination, optimization in every outward facet direction certifies that no feasible grid point violates any such inequality. Hence, no feasible grid point lies outside the returned polytope, and the procedure recovers the convex hull of all feasible grid points without exhaustively enumerating them.

\begin{figure}[t]
    \centering
    \begin{tabular}{@{}c@{\hspace{0.01\linewidth}}c@{\hspace{0.01\linewidth}}c@{\hspace{0.01\linewidth}}c@{}}
    \includegraphics[width=\implementationpanelwidth,trim={95pt 27pt 1126pt 25pt},clip]{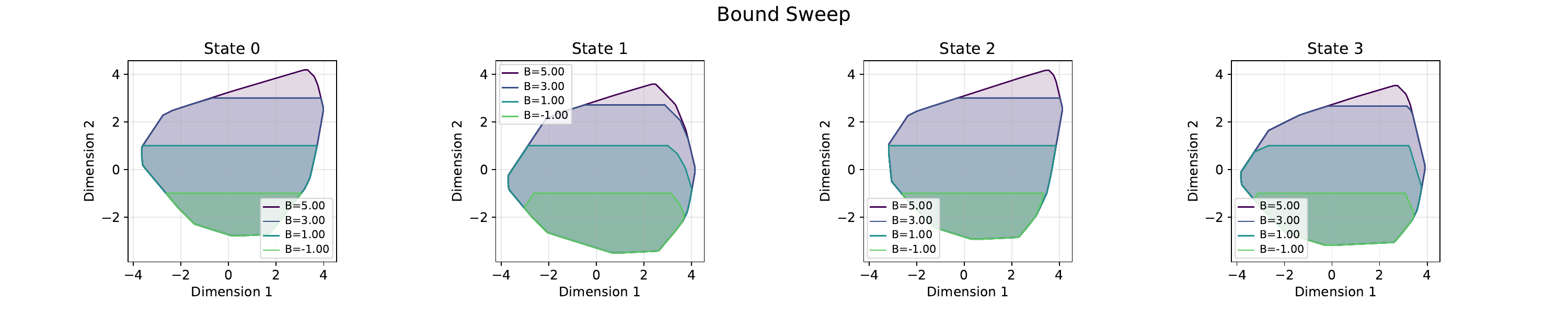} &
    \includegraphics[width=\implementationpanelwidth,trim={433pt 27pt 788pt 25pt},clip]{experiments/bound_sweep.pdf} &
    \includegraphics[width=\implementationpanelwidth,trim={771pt 27pt 450pt 25pt},clip]{experiments/bound_sweep.pdf} &
    \includegraphics[width=\implementationpanelwidth,trim={1109pt 27pt 112pt 25pt},clip]{experiments/bound_sweep.pdf}
    \end{tabular}
    \caption{\small
    Achievable value sets as the cost bound decreases, for the same instance as in \Cref{fig:snapshots}. The horizontal and vertical axes show expected discounted reward and cost, respectively. The cost bound in \Cref{fig:snapshots,fig:trajectory} is $B=2.0$.}
    \label{fig:bound_sweep}
\end{figure}

\subsection{Code}

Further details and a Python implementation of the VSI algorithm are available in the following GitHub repository: \vsirepourl.
Since $\xPhi(\calV)(s)$ can be computed independently for each state $s$, the implementation supports parallel computation across states to improve efficiency.

\Cref{fig:snapshots,fig:trajectory,fig:bound_sweep} present results for an instance with $4$ states and $5$ actions. These results can be reproduced using the code provided in the repository.

\section{Conclusion}

We introduced S-CMDPs, which strengthen classical CMDPs by requiring cost constraints to hold for every subprocess encountered during execution.
Algorithmically, this seemingly small change fundamentally alters the problem: stationary policies can be suboptimal and are computationally intractable. 
We adapted the VSI framework to show that history-dependent policies are surprisingly tractable in this setting.
Besides the theoretical results, we also provide a Python implementation based on MILP vertex computation; details can be found in \Cref{sec:implementation}.

Two natural next steps are to consider multiple cost functions and move beyond the model-based setting. Although our algorithm extends to $d$ cost functions by enlarging the value space from $\mathbb{R}^2$ to $\mathbb{R}^{d+1}$, the number of grid points grows exponentially in $d$. Developing scalable approaches for such settings---potentially combining our structural insights with heuristic or RL-based methods---remains an interesting direction.

\bibliographystyle{plainnat}
\bibliography{ref}

\newpage

\onecolumn

\appendix

\setboolean{isappendix}{true}

\section{Omitted Proofs}

\subsection{Proofs in \Cref{sec:why-not-stationary}}

\thmhardness*

\begin{proof}

\newcommand{\betagapsat}{{\sc GAP 3-SAT}}
\newcommand{\termin}{\perp}

We reduce from the {\betagapsat} problem.
An instance is a 3CNF formula $\varphi$, promised to be either satisfiable (a yes-instance) or such that no Boolean assignment satisfies more than a $\beta$ fraction of its clauses (a no-instance).
This problem is NP-hard for any constant $\beta \in (7/8,1)$ \cite{haastad2001some}.

Consider a {\betagapsat} instance $\varphi$.
Let $x_1,\dots,x_m$ be its Boolean variables and $C_1,\dots,C_n$ its clauses.
Each clause is a disjunction of three literals, each of which is a variable $x_i$ or its negation $\neg x_i$.

We construct an {S-CMDP} instance in which every trajectory reaches an absorbing terminal state within four steps.
The state space is $\{s_0,s_\termin\}\cup\{x_1,\dots,x_m\}\cup\bigcup_{j=1}^n\{C_j,\ell_{j1},\ell_{j2},\ell_{j3}\}$.
An example of the reduction is illustrated in \Cref{fig:reduction}.
In more detail:
\begin{itemize}
\item 
$s_\termin$ is an absorbing terminal state with zero reward and cost.
Each state $x_i$ corresponds to the variable $x_i$ in the {\betagapsat} instance;
each state $C_j$ corresponds to the clause $C_j$; 
and each state $\ell_{jk}$ corresponds to the $k$-th literal in $C_j$.

\item 
The initial state is $s_0$. Irrespective of the action taken, it transitions to each clause state $C_j$ with probability $1/n$, generating zero reward and cost.

All other transitions are deterministic:
\begin{itemize}
\item 
There are four actions available at each $C_j$:
one leading to $s_\termin$, and three others each leading to one of the literal states $\ell_{j1}, \ell_{j2}, \ell_{j3}$.
All these transitions generate zero reward and zero cost; for all $a \in A$:
\begin{align*}
r(C_j, a) = c(C_j, a) = 0.
\end{align*}
(The action leading to $s_\termin$ ensures that there is a feasible policy for every instance.)

\item 
At each literal state $\ell_{jk}$, the only available action leads to the corresponding variable state $x_i$.
Its reward and cost depend on the sign of the literal:
\begin{align*}
r(\ell_{jk}, a) = c(\ell_{jk}, a) = 
\begin{cases}
    \gamma & \text{if $\ell_{jk} = x_i$,}\\
    0 & \text{if $\ell_{jk} = \neg x_i$.}
\end{cases}
% \quad
% \text{for all } a \in A.
\end{align*}

\item 
At each variable state $x_i$, there are two actions available; call them $a_0$ and $a_1$. Both actions lead to $s_\termin$, generating the following reward and cost:
\begin{align*}
& r(x_i, a_0) = c(x_i, a_0) = 0, \\
\text{and} \quad 
& r(x_i, a_1) = c(x_i, a_1) = 1.
\end{align*}
\end{itemize}
Hence, the reward and cost are equal for every state-action pair in the {S-CMDP} instance.

\item 
Finally, we set the thresholds in the constraints as follows.
\begin{align*}
\theta(s) = 
\begin{cases}
\gamma & \text{if $s$ is a positive literal state,} \\
2 & \text{otherwise.}
\end{cases}
\end{align*}
All onward costs are at most $2$, so only the positive literal states have nontrivial constraints.
\end{itemize}
The construction has polynomial size. State-dependent action sets can be replaced by a common action set by duplicating available actions.

\begin{figure}[t]
\centering
\tikzstyle{cstate}=[state, initial, initial where=above, draw=black, text=black]
\tikzstyle{xstate}=[state, draw=black,fill=black!10]
\tikzstyle{lstate}=[state, text=black, minimum size=8mm]
\tikzstyle{minoredge}=[-,thick, dotted]
% \tikzstyle{nolabelstate}=[state, minimum size=4.5mm,text width=0mm]

\begin{tikzpicture}[baseline={(current bounding box.center)}, node distance=27mm, scale=0.8, transform shape]
\tikzstyle{every node}=[inner sep=1pt, align=center, font=\small] 

% \node[state, initial, initial where=above] (sinit) {$s_0$};
% --
\node[cstate] (C1) {$C_1$};
\node[cstate, right of=C1, xshift=5mm] (C2) {$C_2$};
\node[right of=C2, xshift=-4mm] (dots1) {$\dots$};
\node[cstate, right of=dots1, xshift=-4mm] (Cn) {$C_n$};
% --
\node[terminalState, below left of=C1, xshift=-7mm, yshift=0mm] (t0) {};
% --
\node[xstate, below of=C1, yshift=-14mm] (x1) {$x_1$};
\node[xstate, right of=x1] (x2) {$x_2$};
\node[xstate, right of=x2] (x3) {$x_3$};
\node[xstate, right of=x3] (x4) {$x_4$};
\node[lstate] (l11) at ($(C1)!0.5!(x1)$) {$\ell_{11}$};
\node[lstate] (l12) at ($(C1)!0.5!(x2)$) {$\ell_{12}$};
\node[lstate] (l13) at ($(C1)!0.5!(x4)$) {$\ell_{13}$};
\node at ($(l11)+(-1,0)$) {$\theta=\gamma$};
\node at ($(l12)+(1,0)$) {$\theta=\gamma$};
\node at ($(l13)+(1,0)$) {$\theta=2$};
% --
\node[terminalState, below left of=x1, xshift=8mm] (t1) {};
\node[terminalState, below right of=x1, xshift=-8mm] (t2) {};
% ---------------------------
\draw 
% --
(C1) edge[] node[edgeLabel,pos=0.5]{$0$} (t0)
(C1) edge[] node[edgeLabel,pos=0.5]{$0$} (l11)
(C1) edge[] node[edgeLabel,pos=0.5]{$0$} (l12)
(C1) edge[] node[edgeLabel,pos=0.5]{$0$} (l13)
(l11) edge[] node[edgeLabel,pos=0.5]{$\gamma$} (x1)
(l12) edge[] node[edgeLabel,pos=0.5]{$\gamma$} (x2)
(l13) edge[] node[edgeLabel,pos=0.5]{$0$} (x4)
% ---
(C2) edge[minoredge] ($(C2)!0.2!(x1)$)
(C2) edge[minoredge] ($(C2)!0.3!(x2)$)
(C2) edge[minoredge] ($(C2)!0.25!(x3)$)
(C2) edge[minoredge] ($(C2)!0.2!(x4)$)
% --
(Cn) edge[minoredge] ($(Cn)!0.2!(x2)$)
(Cn) edge[minoredge] ($(Cn)!0.3!(x3)$)
(Cn) edge[minoredge] ($(Cn)!0.35!(x4)$)
(Cn) edge[minoredge] ($(Cn)+(0.6,-0.6)$)
% --
(x1) edge[] node[left,pos=0.2,xshift=-1mm]{$a_0$} node[edgeLabel,pos=0.6]{$0$} (t1)
(x1) edge[] node[right,pos=0.2,xshift=1mm]{$a_1$} node[edgeLabel,pos=0.6]{$1$} (t2)
% --
(x2) edge[minoredge] ($(x2)+(-0.4,-0.7)$)
(x2) edge[minoredge] ($(x2)+(0.4,-0.7)$)
(x3) edge[minoredge] ($(x3)+(-0.4,-0.7)$)
(x3) edge[minoredge] ($(x3)+(0.4,-0.7)$)
(x4) edge[minoredge] ($(x4)+(-0.4,-0.7)$)
(x4) edge[minoredge] ($(x4)+(0.4,-0.7)$)
;

\end{tikzpicture}
\vspace{3mm}
\caption{Reduction from {\betagapsat}. In this example, there are four variables $x_1,\dots, x_4$ in $\varphi$ and $C_1 = (x_1 \vee x_2 \vee \neg x_4)$.
The states connected to $C_1$ are shown. The initial state $s_0$, which transitions uniformly to the clause states, is omitted. Edge labels give the reward, which equals the cost in this reduction.
}
\label{fig:reduction}
\end{figure}
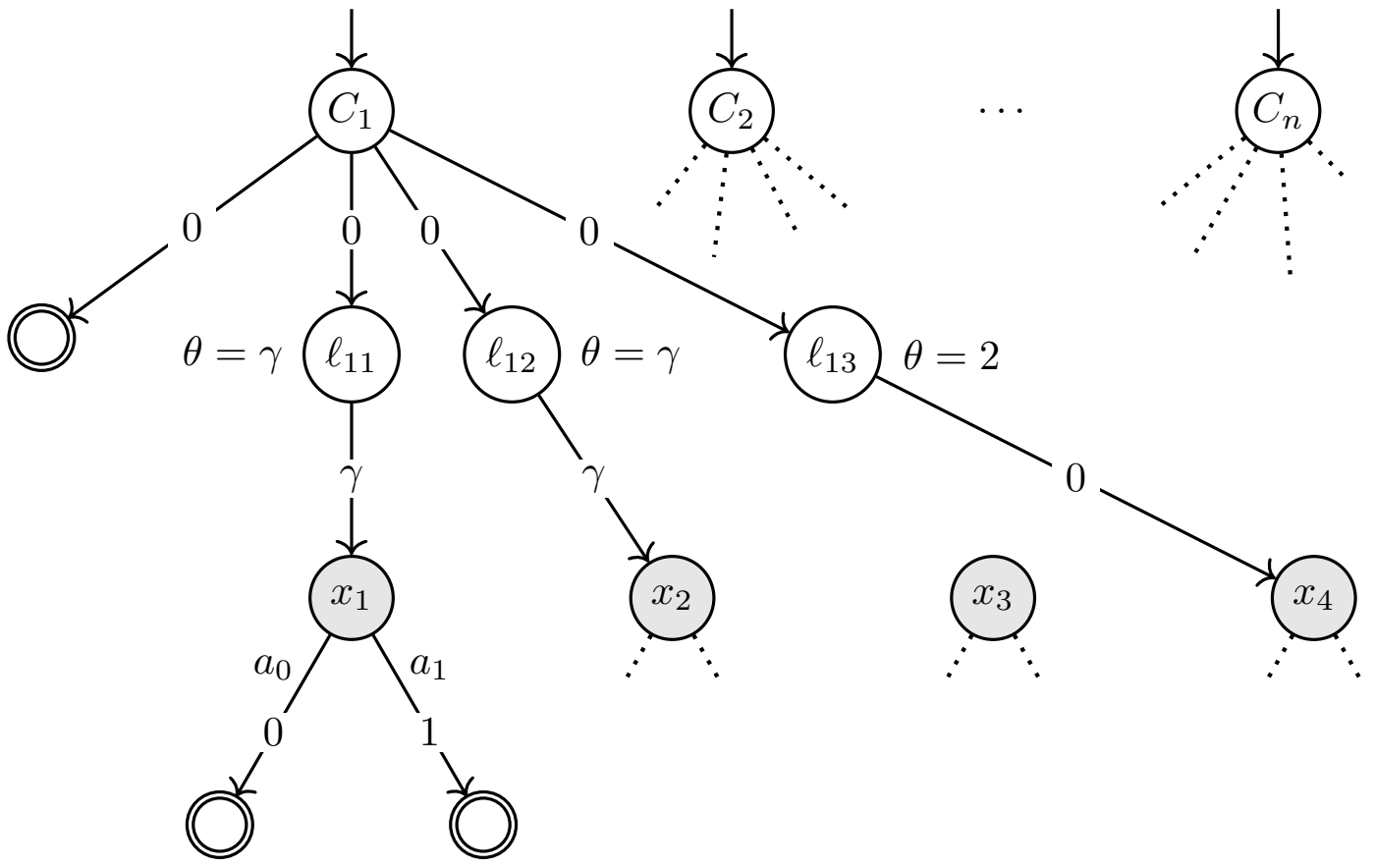

Fix $\beta\in(7/8,1)$, choose $\eta\in(0,1-\beta)$, and set $\delta=\gamma\eta$.
We establish the following two claims:
\begin{itemize}
\item {\bf Completeness:}
If the {\betagapsat} instance is a yes-instance, then there exists a feasible stationary policy $\pi$ with $V_r^\pi(s_0)=\gamma^3$.
\item {\bf Soundness:}
If it is a no-instance, then $V_r^\pi(s_0)\le(\beta+\eta)\gamma^3$ for every $\delta$-feasible stationary policy $\pi$.
\end{itemize}
Taking $\epsilon=(1-\beta-\eta)\gamma^3/2 > 0$ then leaves a strict gap between the value guaranteed in a yes-instance and the upper bound in a no-instance.
Note that it is essential that, in the completeness condition, the yes-instance corresponds to a policy that satisfies the constraints {\em exactly} because of the resource augmentation in the approximation criterion considered (\Cref{def:approx-opt}).

\paragraph{Proof of Completeness}
Suppose that $\varphi$ is a yes-instance, so there exists a satisfying assignment $f: \{x_1,\dots,x_m\} \to \{\true, \false\}$.
Consider the following stationary policy $\pi:S \to \Delta(A)$. We show that $V_r^\pi(s_0) = \gamma^3$.
\begin{itemize}
\item At each variable state $x_i$, choose $a_0$ if $f(x_i)=\true$, and $a_1$ otherwise.

\item At each clause state $C_j$, choose with probability $1$ an action leading to a literal that is true under $f$. Such a literal exists because $f$ satisfies $\varphi$.
\end{itemize}
If the chosen literal is positive, its immediate reward and cost are $\gamma$, and the next action is $a_0$, with zero reward and cost.
If the chosen literal is negative, its immediate reward and cost are zero, and the next action is $a_1$, with reward and cost $1$.
Thus, in both cases the onward reward and cost at the literal state equal $\gamma$. All constraints are satisfied, each clause state has reward value $\gamma^2$, and $V_r^\pi(s_0)=\gamma^3$.

\paragraph{Proof of Soundness}
Suppose that $\varphi$ is a no-instance, and let $\pi$ be any $\delta$-feasible stationary policy.
Write 
\[
    q_i = \pi(a_1 \given x_i).
\]
At a reachable positive literal corresponding to $x_i$, the onward cost is $\gamma + \gamma q_i$. Since its threshold is $\gamma$, $\delta$-feasibility requires $q_i \le \delta / \gamma = \eta$.
Thus, a positive literal with $q_i > \eta$ cannot be reached under $\pi$.

Let $T = \{x_i \mid q_i \le \eta\}$ and let $F$ contain the remaining variables.
We bound the reward at each clause state $C_j$ by considering three cases:
\begin{itemize}
\item {\bf Case 1:} $C_j$ contains a positive literal $x_i$ with $x_i\in T$.
Selecting such a literal gives reward $\gamma^2 (1+q_i) \le (1+\eta) \gamma^2$ at $C_j$. Any selectable positive literal satisfies the same bound, and any negative literal gives at most $\gamma^2$. Thus $V_r^\pi(C_j) \le (1 + \eta) \gamma^2$.

\item {\bf Case 2:} Case~1 does not apply, but $C_j$ contains a negative literal $\neg x_i$ with $x_i\in F$.
All positive literals in the clause are unreachable under $\pi$, and selecting any negative literal gives reward $\gamma^2 q_i \le \gamma^2$ at $C_j$. Hence $V_r^\pi(C_j) \le \gamma^2$.

\item {\bf Case 3:} Neither of the above cases applies.
Every positive literal has its variable in $F$ and is therefore unreachable, while every negative literal has its variable in $T$ and gives reward at most $\eta \gamma^2$ at $C_j$. The terminal action gives zero reward, so $V_r^\pi(C_j) \le \eta \gamma^2$.
\end{itemize}

Assigning $\true$ to the variables in $T$ and $\false$ to those in $F$ satisfies every clause in Cases~1 and~2. Since $\varphi$ is a no-instance, these cases cover at most a $\beta$ fraction of the clauses.
Their reward values are at most $(1 + \eta) \gamma^2$, while the remaining clause values are at most $\eta \gamma^2$.
Accounting for the initial transition from $s_0$, we obtain
\[
V_r^\pi(s_0)
\;=\; \frac{\gamma}{n} \sum_{j=1}^n V_r^\pi(C_j)
\;\le\; \gamma \bigl( \beta (1 + \eta) \gamma^2 + (1 - \beta) \eta \gamma^2 \bigr)
\;=\; ( \beta + \eta ) \gamma^3.
\]

\medskip

In a yes-instance, any $(\epsilon,\delta)$-optimal stationary policy has value at least $\gamma^3-\epsilon$. In a no-instance, every $\delta$-feasible stationary policy has value at most $(\beta+\eta)\gamma^3$. Since $\gamma^3-\epsilon>(\beta+\eta)\gamma^3$, computing such a policy would distinguish the two cases, proving the claimed hardness. Since all nonzero rewards and costs occur within four steps, the result also holds for a finite horizon.
\end{proof}

\subsection{Proofs in \Cref{sec:vsi}}

\lmmcharacterization*

\begin{proof}
The ``only if'' direction follows directly from the Bellman equation and the subprocess feasibility requirement.

For the ``if'' direction, suppose $v_2 \le \theta(s)$ and there exist $\barpi$ and $\ww$ satisfying the stated conditions.
By \Cref{eq:const-onward}, for each pair $(a,s')$ with $\barpi(a) \cdot p(s'\given s,a)>0$, there exists a feasible policy $\pi_{a,s'}$ that achieves the onward vector $\ww(a, s')$ from state $s'$.
We construct a policy $\pi$ by concatenating $\barpi$ and these policies: $\pi$ selects its initial action at $s$ according to $\barpi$, and thereafter follows $\pi_{a,s'}$ once action $a$ is taken and state $s'$ is reached. Its action distributions on unreachable histories can be chosen arbitrarily.
Since $v_2 \le \theta(s)$ and each $\pi_{a,s'}$ is feasible, the resulting policy $\pi$ satisfies the subprocess constraints. Applying the Bellman equation then yields $\VV^\pi(s)=\vv$. Hence $\vv$ is achievable: $\vv \in \calV^\star(s)$.
\end{proof}

\lmmmonotonicity*

\begin{proof}
It is immediate from the definition that $\onward(s, \cdot)$ is monotone. Hence, if $\calV \subseteq \calV'$, then we have $\onward(s, \calV) \subseteq \onward(s, \calV')$. 
According to the definition of $\Phi$, enlarging the feasible set $\onward(s, \calV)$ can only enlarge the resulting value set. Therefore, $\Phi(\calV)(s) \subseteq \Phi(\calV')(s)$ for all $s \in S$. 
\end{proof}

\lmmballproperties*

\begin{proof}
The inclusion $\calV^\star \subseteq \calB$ follows from the boundedness of discounted returns. Since rewards and costs are assumed to lie in $[-1,1]$, the discounted cumulative return of any policy is bounded in magnitude by the geometric sum
$\sum_{t=0}^{\infty} \gamma^t = 1/(1-\gamma)$.
Hence every achievable value vector lies in $\calB$.

For the second inclusion, consider any $\vv \in \Phi(\calB)(s)$. By definition of $\Phi$, $\vv$ is obtained from an immediate payoff of magnitude at most $1$ together with an onward value bounded by $1/(1-\gamma)$. Therefore
\[
\|\vv\|_\infty 
\,\le\, 1 + \gamma \cdot \frac{1}{1-\gamma}
\,=\, \frac{1}{1-\gamma},
\]
which implies $\vv \in \calB$. Hence $\Phi(\calB) \subseteq \calB$.
\end{proof}

\lmmnested*

\begin{proof}
We first prove the base case $\calV^\star \subseteq \calV^1 \subseteq \calV^0$.

Since $\calV^\star \subseteq \calB$ (\Cref{lmm:ball-properties}), monotonicity of $\Phi$ implies 
\[
\Phi(\calV^\star) \subseteq \Phi(\calB).
\]

Recall that $\calV^\star = \Phi(\calV^\star)$ (\Cref{crl:fixed-point}). Moreover, by definition of the sequence $(\calV^k)_k$, we have
$\calV^0 = \calB$ and $\calV^1 = \Phi(\calV^0)$.
Combining these results establishes
\[
\calV^\star
= \Phi(\calV^\star)
\subseteq \Phi(\calB)
= \calV^1 .
\]
Together with $\Phi(\calB) \subseteq \calB$ (\Cref{lmm:ball-properties}), this yields the base case
$\calV^\star \subseteq \calV^1 \subseteq \calV^0$.

Now suppose that $\calV^\star \subseteq \calV^{k+1} \subseteq \calV^k$ holds for some $k \ge 0$.
Applying $\Phi$ to the inclusions and using monotonicity gives
\[
\Phi(\calV^\star) \subseteq \Phi(\calV^{k+1}) \subseteq \Phi(\calV^k).
\]
By definition of the sequence $(\calV^k)_k$ and the fact that $\calV^\star = \Phi(\calV^\star)$, this becomes
$\calV^\star \subseteq \calV^{k+2} \subseteq \calV^{k+1}$.

By induction, the stated result then follows.
\end{proof}

\lmmgreatest*

\begin{proof}
Suppose $\calV$ is bounded and $\calV = \Phi(\calV)$. Fix $s \in S$ and $\vv \in \calV(s)$.
We construct a policy $\pi$ and show that it is feasible and satisfies $\VV^\pi(s)=\vv$.

\paragraph{Recursive Policy Construction}
For a history $h$, let $s_h$ denote its last state. We define the policy $\pi$ and targets $\vv_h\in\calV(s_h)$ recursively over reachable histories:
\begin{itemize}
\item
At the initial history $h_0=(s)$, assign the target $\vv_{h_0}=\vv$.

\item
At each reachable history $h$ with target $\vv_h$, choose $(\barpi_h,\ww_h)\in\onward(s_h,\calV)$ such that
\[
\bellman(s_h,\barpi_h,\ww_h)=\vv_h.
\]
Such a pair exists because $\calV=\Phi(\calV)$. Set 
\[
\pi(h)=\barpi_h.
\]

\item
For every pair $(a,s')$ with $\barpi_h(a)\cdot p(s'\given s_h,a)>0$, assign to the reachable extension $h'=(h,a,s')$ the target 
\[
    \vv_{h'}=\ww_h(a,s').
\]
Crucially, the onward condition imposed on $\ww_h$ ensures that $\vv_{h'} \in \calV(s')$, so the construction can continue at $h'$.
\end{itemize}
For completeness, for all unreachable histories $h$, we assign an arbitrary action distribution to $\pi(h)$.

\paragraph{Exact Realization}
Let $\VV^\pi(h)$ denote the true onward value of $\pi$ conditional on $h$. Its Bellman expansion and the exact decomposition of $\vv_h$ use the same immediate reward and cost terms. Subtracting them and using the successor targets gives
\begin{align*}
    \VV^\pi(h)-\vv_h
    &=\gamma\,\E_\pi\!\left[
        \VV^\pi(h')-\vv_{h'}\,\middle|\,h
    \right],\\
    \|\VV^\pi(h)-\vv_h\|_\infty
    &\le\gamma\,\E_\pi\!\left[
        \|\VV^\pi(h')-\vv_{h'}\|_\infty\,\middle|\,h
    \right],
\end{align*}
where $h'$ is the next history obtained from $h$ under $\pi$ and $p$, and the inequality follows from the triangle inequality. Since $\calV$ is bounded, let $M<\infty$ bound the $L_\infty$ norms of its vectors. Every target then has norm at most $M$, while every true onward value has norm at most $1/(1-\gamma)$. Iterating the inequality for $T\ge1$ steps therefore yields
\[
    \|\VV^\pi(h)-\vv_h\|_\infty
    \le \gamma^T\left(\frac{1}{1-\gamma}+M\right).
\]
Letting $T\to\infty$ gives $\VV^\pi(h)=\vv_h$ at every reachable history. In particular, $\VV^\pi(s)=\vv$.

\paragraph{Feasibility}
At every reachable history $h$, we have $\vv_h\in\calV(s_h)=\Phi(\calV)(s_h)$, so $v_{h,2}\le\theta(s_h)$. Since $\VV^\pi(h)=\vv_h$, the policy $\pi$ is feasible. Thus $\vv\in\calV^\star(s)$, proving $\calV\subseteq\calV^\star$.
\end{proof}

\thmconvergence* 

\begin{proof}
Let $\calV^\infty(s) := \bigcap_{k\ge 0} \calV^k(s)$ for every $s \in S$.
By \Cref{lmm:nested}, we already have $\calV^\star \subseteq \calV^\infty\subseteq\calB$. Thus, by \Cref{lmm:greatest}, it suffices to show that $\calV^\infty$ is a fixed point of $\Phi$. We prove the two inclusions separately.

\paragraph{The Inclusion $\calV^\infty\subseteq\Phi(\calV^\infty)$}
We construct a Bellman decomposition in $\calV^\infty$ by taking a limit of decompositions in the iterates.
Fix an arbitrary state $s\in S$ and vector $\vv \in \calV^\infty(s)$. By definition, $\vv \in \calV^k(s)$ for all $k \ge 0$. 
Since $\calV^{k+1} = \Phi(\calV^k)$, we also have $\vv \in \Phi(\calV^k)(s)$. 
According to the definition of $\Phi$, for each $k$, there exists  $(\bar{\pi}^k, \ww^k) \in \onward(s, \calV^k)$ such that
\[
\vv = \bellman(s,\bar{\pi}^k, \ww^k).
\]

Choose $\ww^k(a,s')=\mathbf{0}$ whenever $\barpi^k(a) \cdot p(s'\given s,a)=0$; this does not affect the Bellman expansion. All onward vectors then lie in $\calB$.
Consider the sequence $\left( \bar{\pi}^k, \ww^k \right)_{k \ge 0}$.
Since both components are bounded, by the Bolzano--Weierstrass theorem,
this sequence admits a convergent subsequence $\left(\bar{\pi}^{k_\ell}, \ww^{k_\ell}\right)_{\ell \ge 0}$. 
Let $(\bar{\pi}, \ww)$ be the limit of this subsequence.

The key is to show that the limiting onward vectors for positive-probability pairs remain in every iterate, and hence in their intersection.

\begin{claim}
\label{clm:1}
$(\bar{\pi}, \ww) \in \onward(s, \calV^\infty)$.
\end{claim}

\begin{proof}
According to the definition of $\onward(s, \cdot)$ in \Cref{eq:onward}, we need to show that: 
\begin{itemize}
\item[(i)] $\bar{\pi} \in \Delta(A)$; and 
\item[(ii)] $\ww(a, s') \in \calV^\infty(s')$ for all $a$ and $s'$ such that $\barpi(a) \cdot p(s' \given s, a) > 0$.
\end{itemize}

For (i), note that $\Delta(A)$ is closed and each $\bar{\pi}^{k_\ell} \in \Delta(A)$. Since $\barpi$ is their limit, it follows that $\bar{\pi} \in \Delta(A)$. 

For (ii), fix $a \in A$ and $s' \in S$ such that $\barpi(a) \cdot p(s' \given s, a) > 0$. 
Let $k \ge 0$ be arbitrary. Since $k_\ell \to \infty$ and $\barpi^{k_\ell}(a)\to\barpi(a)>0$, there exists $L$ such that $k_\ell \ge k$ and $\barpi^{k_\ell}(a) \cdot p(s'\given s,a)>0$ for all $\ell \ge L$.
Because $(\barpi^{k_\ell}, \ww^{k_\ell}) \in \onward(s, \calV^{k_\ell})$ we have 
\[
\ww^{k_\ell}(a, s') \;\in\; \calV^{k_\ell}(s') \;\subseteq\; \calV^{k}(s')
\]
for all $\ell \ge L$.

To pass to the limit, we use the fact that $\Phi$ preserves closedness of bounded closed value sets. Indeed, all onward vectors can be restricted to a closed ball containing the input sets. The onward constraints remain valid under limits: any action--state pair with positive limiting probability has positive probability for all sufficiently large indices, so its onward vectors lie in the corresponding closed input set, which also contains their limit. Together with the closed cost constraint, this makes the feasible decompositions a compact set. Its continuous Bellman image is compact and hence closed. Since $\calV^0(s')=\calB$ is closed and all iterates are bounded, induction shows that every $\calV^k(s')$ is closed.

Taking limits therefore gives $\ww(a, s') \in \calV^k(s')$.
Since $k$ is arbitrary, we conclude that $\ww(a,s') \in \bigcap_{k\ge 0} \calV^k(s') = \calV^\infty(s')$.
\end{proof}

We can now pass to the limit in the Bellman expansion.
Since $\vv = \bellman(s, \bar{\pi}^{k_\ell}, \ww^{k_\ell})$ for all $\ell$, and $\bellman(s,\cdot,\cdot)$ is continuous, taking limits along the subsequence yields
$\vv = \bellman(s, \bar{\pi}, \ww)$.
Moreover, $v_2\le\theta(s)$ since $\vv\in\calV^1(s)$. Together with \Cref{clm:1}, this implies $\vv \in \Phi(\calV^\infty)(s)$.
Because the choice of $s \in S$ and $\vv \in \calV^\infty(s)$ is arbitrary, we conclude that
\[
\calV^\infty \subseteq \Phi(\calV^\infty).
\]

\paragraph{The Reverse Inclusion}
Since $\calV^\infty\subseteq\calV^k$ for every $k\ge0$, monotonicity gives $\Phi(\calV^\infty)\subseteq\Phi(\calV^k)=\calV^{k+1}$. Taking intersections yields, for every $s\in S$,
\[
    \Phi(\calV^\infty)(s)
    \subseteq \bigcap_{k\ge0}\calV^{k+1}(s)
    = \bigcap_{k\ge0}\calV^k(s)
    = \calV^\infty(s),
\]
where dropping $\calV^0(s)$ does not change the intersection because the iterates are nested.

Therefore, $\calV^\infty=\Phi(\calV^\infty)$. By \Cref{lmm:greatest}, $\calV^\infty\subseteq\calV^\star$, and the reverse inclusion from \Cref{lmm:nested} gives $\calV^\infty=\calV^\star$.
\end{proof}

\subsection{Proofs in \Cref{sec:apprx-vsi}}

\lmmnestedx*

% \begin{proof}
% It suffices to verify that conditions analogous to those in \Cref{lmm:ball-properties} hold for the operator $\xPhi$, so that the same argument used in the proof of \Cref{lmm:nested} applies verbatim.

% In particular, we need to establish:
% (i) $\calV^\star \subseteq \xcalV^0$, and  
% (ii) $\xPhi(\xcalV^0) \subseteq \xcalV^0$.

% Condition (i) is immediate: $\calV^\star \subseteq \calB \subseteq \neigh_\xi(\calB) = \xcalV^0$.

% To verify (ii), observe that
% \[
% \Phi \left(\neigh_\xi(\calB) \right) \subsetneq \neigh_\xi(\calB).
% \]
% This follows from the bound
% \[
% 1 + \gamma\!\left(\frac{1}{1-\gamma} + \xi \right)
% \;=\; \frac{1}{1-\gamma} + \gamma \xi
% \;<\; \frac{1}{1-\gamma} + \xi,
% \]
% which holds since $\gamma < 1$.
% In other words, when the immediate return is bounded by $1$ and the onward value is bounded by $\frac{1}{1-\gamma} + \xi$ (the radius of $\neigh_\xi(\calB)$), the resulting discounted cumulative return remains strictly within $\neigh_\xi(\calB)$.

% Recall that $\xi$ is chosen to be a constant that divides $1/(1-\gamma)$.
% In this case, $\Phi \left(\neigh_\xi(\calB) \right) \subsetneq \neigh_\xi(\calB)$ implies that $\Phi \left(\neigh_\xi(\calB) \right)$ does not contain more grid points than does $\calB$.
% By definition of $\xPhi$, we then have $\xPhi(\calB) \subseteq \calB$.
% Since $\xcalV^0 = \calB$, (ii) then follows.
% \end{proof}

\begin{proof}
Let $r_\xi$ denote the radius of $\xcalB_\xi$. By definition,
\[
    r_\xi > \frac{1}{1-\gamma}
    \qquad\text{and}\qquad
    r_\xi \in \xi\mathbb{N},
\]
and $\xcalV^0 = \xcalB_\xi$.

We first establish
(i) $\calV^\star \subseteq \xcalV^0$, and
(ii) $\xPhi(\xcalV^0) \subseteq \xcalV^0$.
Indeed, condition (i) is immediate, since
\[
    \calV^\star
    \subseteq
    \calB
    \subseteq
    \xcalB_\xi
    =
    \xcalV^0.
\]

To verify (ii), fix any $s \in S$ and $\vv \in \Phi(\xcalV^0)(s)$.
Since the immediate return is bounded in $L_\infty$ norm by $1$ and every onward value in $\xcalV^0$ has $L_\infty$ norm at most $r_\xi$, we get that
\[
    \|\vv\|_\infty
    \le
    1 + \gamma r_\xi
    <
    r_\xi,
\]
where the strict inequality follows from
$r_\xi > 1/(1-\gamma)$.

Now consider any
\[
    \mathbf{g} \in
    \neigh_\xi\!\left(\Phi(\xcalV^0)(s)\right)
    \cap G_\xi.
\]
There exists $\vv \in \Phi(\xcalV^0)(s)$ such that 
$\|\mathbf{g} - \vv\|_\infty \le \xi$, and hence
$\| \mathbf{g} \|_\infty < r_\xi + \xi$.
Since $\mathbf{g} \in G_\xi$, $\| \mathbf{g} \|_\infty$ is a multiple of $\xi$.
As $r_\xi$ is also a multiple of $\xi$, it follows that
\[
    \| \mathbf{g} \|_\infty \le r_\xi.
\]
Thus every such grid point belongs to $\xcalV^0$. Since
$\xcalV^0$ is convex, by the definition of $\xPhi$, we get that
\[
    \xPhi(\xcalV^0)(s)
    \;=\;
    \gridhull_\xi\!\left(
        \neigh_\xi\!\left(\Phi(\xcalV^0)(s)\right)
    \right)
    \;\subseteq\;
    \xcalV^0.
\]
Since $s$ was arbitrary, condition (ii) follows.

\smallskip

Finally, observe that for any value-set function $\calV$, we have
\[
    \Phi(\calV)(s)
    \subseteq
    \xPhi(\calV)(s)
    \qquad\text{for every } s \in S.
\]
Indeed, fix any $\vv \in \Phi(\calV)(s)$ and consider a grid cell
containing $\vv$. Every vertex $\mathbf{g}$ of this cell belongs to
$G_\xi$ and satisfies
$\| \mathbf{g} -\vv\|_\infty \le \xi$,
so
\[
    \mathbf{g} \in
    \neigh_\xi\!\left(\Phi(\calV)(s)\right)
    \cap G_\xi.
\]
Since $\vv$ is a convex combination of the vertices of the cell,
the definition of $\gridhull_\xi$ then implies
$\vv \in \xPhi(\calV)(s)$;
therefore, $\Phi(\calV)(s) \subseteq \xPhi(\calV)(s)$.

\smallskip

Applying this to $\calV^\star$ and using
$\calV^\star=\Phi(\calV^\star)$ gives
\[
    \calV^\star
    =
    \Phi(\calV^\star)
    \subseteq
    \xPhi(\calV^\star).
\]
Together with the monotonicity of $\xPhi$, conditions (i) and (ii)
give
\[
    \calV^\star
    \subseteq
    \xcalV^1
    \subseteq
    \xcalV^0.
\]
Applying $\xPhi$ repeatedly and using
$\calV^\star \subseteq \xPhi(\calV^\star)$ then yields
\[
    \calV^\star
    \subseteq
    \xcalV^{k+1}
    \subseteq
    \xcalV^k
\]
for every $k \ge 0$.
\end{proof}

\lmmepsdeltaachievable*

\begin{proof}
Suppose $\calV$ is bounded and $\calV = \xPhi(\calV)$. Fix $s \in S$ and $\vv \in \calV(s)$.
We construct a policy $\pi$ and show that it is $\delta$-feasible and satisfies $\|\VV^\pi(s)-\vv\|_\infty\le\epsilon$. The construction uses a relaxed version of $\Achieve$ (\Cref{alg:achieve}), allowing a Bellman error of at most $\xi$ at each step.

\paragraph{Recursive Policy Construction}
We first note that, for every $s'\in S$,
\[
    \xPhi(\calV)(s')
    = \gridhull_\xi\!\left(\neigh_\xi\!\left(\Phi(\calV)(s')\right)\right)
    \subseteq \neigh_\xi\!\left(\Phi(\calV)(s')\right).
\]
Indeed, $\calV=\xPhi(\calV)$ has convex values, so $\Phi(\calV)(s')$ is convex by mixing Bellman decompositions. Its $\xi$-neighborhood is therefore convex and contains the convex hull of its grid points. Using $\calV=\xPhi(\calV)$ now gives
\[
    \calV(s')\subseteq\neigh_\xi\!\left(\Phi(\calV)(s')\right).
\]

For a history $h$, let $s_h$ denote its last state. We define the policy $\pi$ and targets $\vv_h\in\calV(s_h)$ recursively over reachable histories:
\begin{itemize}
\item 
At the initial history $h_0=(s)$, assign the target $\vv_{h_0}=\vv$.

\item 
At each reachable history $h$ with target $\vv_h$, choose $(\barpi_h,\ww_h)\in\onward(s_h,\calV)$ such that
\begin{equation}
\label{eq:selection-barpi-ww}
    \bellman(s_h,\barpi_h,\ww_h) \in \neigh_\xi(\vv_h),
\end{equation}
and set 
\[
    \pi(h)=\barpi_h.
\]
Such a choice exists because the inclusion above ensures that some vector in $\Phi(\calV)(s_h)$ lies within distance $\xi$ of $\vv_h$. By the definition of $\Phi$, that vector admits a Bellman decomposition using a pair in $\onward(s_h,\calV)$, which satisfies \Cref{eq:selection-barpi-ww}.

\item 
For every pair $(a,s')$ with $\barpi_h(a)\cdot p(s'\given s_h,a)>0$, assign to the reachable extension $h'=(h,a,s')$ the target
\[
    \vv_{h'} = \ww_h(a,s') \in \calV(s').
\]
The membership $\ww_h(a,s') \in \calV(s')$ follows from the onward condition; it ensures that the recursive construction can continue at $h'$.
\end{itemize}
Define $\pi$ arbitrarily on unreachable histories. The construction uses precisely the relaxed expansion in \Cref{alg:expand-x}, with $\calV$ in place of $\xcalV^\star$.

\paragraph{Approximation Error}
For each reachable history $h$, write $\tilde{\vv}_h=\bellman(s_h,\barpi_h,\ww_h)$. By \Cref{eq:selection-barpi-ww},
\begin{equation}
\label{eq:local-err}
    \|\tilde{\vv}_h-\vv_h\|_\infty \le \xi.
\end{equation}
Let $\VV^\pi(h)$ denote the true onward value of $\pi$ conditional on $h$. Define the onward-value function $\vec{W}_h$ by
\[
    \vec{W}_h(a,s')=\VV^\pi(h,a,s')
\]
for each reachable extension $(h,a,s')$, assigning arbitrary values on zero-probability pairs. The Bellman equation then gives
\[
    \VV^\pi(h)=\bellman(s_h,\barpi_h,\vec{W}_h),
    \qquad
    \tilde{\vv}_h=\bellman(s_h,\barpi_h,\ww_h).
\]
Both expressions use the same immediate reward and cost terms. Subtracting them and using the successor targets gives
\[
    \VV^\pi(h)-\tilde{\vv}_h
    =\gamma\,\E_\pi\!\left[
        \VV^\pi(h')-\vv_{h'}
        \,\middle|\,h
    \right],
\]
where $h'$ is the next history obtained from $h$ under $\pi$ and $p$.
Taking the $L_\infty$ norm and applying the triangle inequality and \Cref{eq:local-err} gives
\begin{align*}
    \|\VV^\pi(h)-\vv_h\|_\infty
    &\le \|\VV^\pi(h)-\tilde{\vv}_h\|_\infty
        +\|\tilde{\vv}_h-\vv_h\|_\infty\\
    &\le \xi+\gamma\,\E_\pi\!\left[
        \|\VV^\pi(h')-\vv_{h'}\|_\infty
        \,\middle|\,h
    \right]\\
    &\le \xi+\gamma\max_{h'}
        \|\VV^\pi(h')-\vv_{h'}\|_\infty.
\end{align*}
Here, the expectation is a weighted average over all histories $h'$ reachable from $h$ in one step, so it is bounded by the maximum over those histories.
Since $\calV$ is bounded, let $M<\infty$ bound the $L_\infty$ norms of its vectors. Every target then has norm at most $M$, while every true onward value has norm at most $1/(1-\gamma)$. Iterating this inequality for $T\ge 1$ steps therefore yields
\[
    \|\VV^\pi(h)-\vv_h\|_\infty
    \le \xi\sum_{j=0}^{T-1}\gamma^j
    +\gamma^T\left(\frac{1}{1-\gamma}+M\right).
\]
Letting $T\to\infty$ gives
\begin{equation}
\label{eq:global-err}
    \|\VV^\pi(h)-\vv_h\|_\infty
    \le \frac{\xi}{1-\gamma} \le \epsilon.
\end{equation}
Applying this at $h_0=(s)$ gives $\|\VV^\pi(s)-\vv\|_\infty\le\epsilon$.

\paragraph{Feasibility}
Every target $\vv_h$ lies within distance $\xi$ of $\Phi(\calV)(s_h)$, whose vectors satisfy the cost threshold at $s_h$. Hence
\begin{equation}
\label{eq:target-feasibility}
    v_{h,2}\le\theta(s_h)+\xi.
\end{equation}
Together with \Cref{eq:global-err}, this yields, at every reachable history $h$,
\[
    V^\pi_2(h)
    \le v_{h,2}+\frac{\xi}{1-\gamma}
    \le \theta(s_h)+\xi+\frac{\xi}{1-\gamma}
    \le \theta(s_h)+\delta,
\]
where the final inequality follows from $\xi\le(1-\gamma)\min\{\epsilon,\delta/2\}$.
Thus $\pi$ is $\delta$-feasible and $\vv$ is $(\epsilon,\delta)$-achievable from $s$.
\end{proof}

\lmmlinearize*

\begin{proof}
The constraints involve two difficulties. 
First, \Cref{eq:computing-bellman} involves quadratic terms of the form $\barpi(a)\cdot \ww(a,s')$. 
Second, the index selection condition in \Cref{eq:computing-onward} depends on the variable $\barpi(a)$. 
We show that both issues can be eliminated to obtain an equivalent formulation with only linear constraints.

\paragraph{Removing the Index Condition}
Observe that \Cref{eq:computing-onward} is equivalent to imposing the following conditions for every $a\in A$ and $s'\in S$, where we write $\barpi(a) \cdot \calV(s') = \left\{ \barpi(a) \cdot \vv \,\middle|\, \vv \in \calV(s') \right\}$:
\[
\begin{aligned}
&
\barpi(a) \cdot \ww(a,s') \in \barpi(a) \cdot \calV(s') &&\qquad \text{if } p(s'\given s,a)>0 \text{ and } \calV(s')\neq\emptyset, \\
&
\barpi(a) = 0 &&\qquad \text{if } p(s'\given s,a)>0 \text{ and } \calV(s')=\emptyset.
\end{aligned}
\]
The first constraint holds automatically if $\barpi(a) = 0$; and if $\barpi(a) > 0$, it reduces to $\ww(a,s') \in \calV(s')$.
Thus the dependence of the index condition on $\barpi(a)$ is removed.

\paragraph{Linearizing Quadratic Terms}

Let $G(s')$ be the finite set of grid points representing $\calV(s')$, so that $\calV(s')=\conv(G(s'))$.
We introduce convex-weight variables only for pairs $(a,s')$ with $p(s'\given s,a)>0$ and $G(s')\neq\emptyset$; empty successor sets with positive transition probability are handled by the zero-action-probability condition above.
Then any $\barpi(a) \cdot \ww(a,s') \in \barpi(a) \cdot \calV(s')$ can be written as a convex combination
\[
\barpi(a) \cdot \ww(a,s') 
= 
\sum_{\mathbf{g} \in G(s')} y_{a,s'}(\mathbf{g}) \cdot \big(\barpi(a) \cdot \mathbf{g} \big),
\]
where $\yy_{a,s'} = (y_{a,s'}(\mathbf{g}))_{\mathbf{g} \in G(s')}$ satisfies
\begin{align}
\label{eq:yy-in}
\yy_{a,s'} \in \Delta(G(s')).
\end{align}
Namely, $\barpi(a) \cdot \calV(s')$ is the convex hull of the grid points scaled by $\barpi(a)$.

Substituting this representation of $\barpi(a) \cdot \ww(a,s')$ into \Cref{eq:computing-bellman} and the reformulation of \Cref{eq:computing-onward} above, we convert the quadratic terms to $\barpi(a) \cdot y_{a,s'}(\mathbf g)$. 
To eliminate these terms, introduce auxiliary variables
\[
z_{a,s'}(\mathbf{g})=\barpi(a) \cdot y_{a,s'}(\mathbf{g}).
\]
Since $\yy_{a,s'} \in \Delta(G(s'))$, the variables $\zz_{a,s'}=(z_{a,s'}(\mathbf{g}))_{\mathbf{g} \in G(s')}$ satisfy the following constraints:
\begin{align}
\label{eq:zz}
z_{a,s'}(\mathbf g)\ge 0,
\qquad
\sum_{\mathbf{g} \in G(s')} z_{a,s'}(\mathbf{g})=\barpi(a).
\end{align}
We replace \Cref{eq:yy-in} with them.

This substitution replaces every product $\barpi(a) \cdot y_{a,s'}(\mathbf g)$ with a linear variable.
The resulting constraints are linear in $\barpi$ and the auxiliary variables $\zz_{a,s'}$.

\paragraph{Equivalence of the Reformulation}

The reformulation is trivially equivalent before introducing $\zz$.
To see that it remains equivalent after we replace $\yy$ with $\zz$, observe that if $(\tilde{\vv}, \barpi, \yy)$ satisfies the original formulation, then $(\tilde{\vv}, \barpi, \zz)$ with 
$\zz_{a,s'} = \barpi(a) \yy_{a,s'}$ satisfies the reformulated constraints. 

Conversely, given feasible $(\tilde{\vv}, \barpi, \zz)$, define, for each pair with $p(s'\given s,a)>0$ and $G(s')\neq\emptyset$,
\[
\yy_{a,s'} = 
\begin{cases}
\zz_{a,s'} / \barpi(a) & \text{if }  \barpi(a) > 0, \\
\text{an arbitrary vector in } \Delta(G(s')) & \text{if }  \barpi(a) = 0.
\end{cases}
\]
Then $(\tilde{\vv}, \barpi, \yy)$ satisfies the original formulation.
In particular, in the case where $\barpi(a) = 0$, \Cref{eq:zz} implies that $\zz_{a,s'} = \mathbf{0}$; hence, any arbitrary $\yy_{a,s'} \in \Delta(G(s'))$ gives $\zz_{a,s'} = \barpi(a) \cdot \yy_{a,s'}$.

\paragraph{The Full LP Reformulation}
In summary, the full LP reformulation is as follows. The variables are:
\begin{itemize}
\item 
$\tilde{\vv} = (\tilde{v}_1, \tilde{v}_2)$;

\item 
$\barpi(a)$ for each $a \in A$; and

\item 
$z_{a,s'}(\mathbf g)$ for each $a\in A$, $s'\in S$, and $\mathbf g\in G(s')$.
\end{itemize}
For uniform notation, we retain the variables for pairs with $p(s'\given s,a)=0$; they are unconstrained and contribute nothing to the Bellman expression.
The constraints are:
\begin{itemize}
\item \Cref{eq:computing-dist,eq:computing-threshold}:
$\|\vv-\tilde{\vv}\|_\infty \le \xi$, and $\tilde{v}_2 \le \theta(s)$.

\item 
The linearized form of \Cref{eq:computing-bellman}, as follows:
\[
\tilde{\vv} \;=\;
        \sum_{a \in A} \barpi(a) \cdot \xx(s, a)
        \;+\;
        \gamma 
        \sum_{a \in A} \sum_{s' \in S} p(s' \given s, a) 
        \sum_{\mathbf{g} \in G(s')} z_{a,s'}(\mathbf{g}) \cdot \mathbf{g}
\]

\item 
The linearized form of \Cref{eq:computing-onward}, for each $a\in A$ and $s'\in S$ such that $p(s'\given s,a)>0$:
\begin{itemize}
    \item 
    If $\calV(s') \neq \emptyset$, then $z_{a,s'}(\mathbf g)\ge 0$ for all $\mathbf{g} \in G(s')$ and $\sum_{\mathbf g\in G(s')}z_{a,s'}(\mathbf g)=\barpi(a)$;

    \item 
    If $\calV(s') = \emptyset$, then $\barpi(a)=0$.
\end{itemize}

\item 
Finally, $\barpi \in \Delta(A)$ so that $\barpi$ is a valid distribution.
\end{itemize}

Any feasible $(\tilde{\vv}, \barpi,\zz)$ yields a feasible solution $(\tilde{\vv}, \barpi,\ww)$ of the original constraints by setting
\[
\ww(a,s')
=
\begin{cases}
\sum_{\mathbf{g} \in G(s')} z_{a,s'}(\mathbf{g})\cdot \mathbf{g} / \barpi(a)  & \text{if } \barpi(a) > 0, \\
\text{arbitrary} & \text{if } \barpi(a) = 0.
\end{cases}
\]

This completes the proof.
\end{proof}

\thmpolytime* 

\begin{proof}
Let $N_\xi=|G_\xi\cap\xcalB_\xi|$. Each application of $\xPhi$ requires at most $|S|N_\xi$ LP feasibility checks: for each state $s$ and grid point $\mathbf{g} \in G_\xi \cap \xcalB_\xi$, we test whether
\[
    \mathbf{g} \in \neigh_\xi(\Phi(\calV)(s)),
\]
using \Cref{lmm:linearize}. The resulting sets can be represented by all accepted grid points, so each LP has polynomial size in the input and $N_\xi$.
By \Cref{lmm:nested-x}, the sequence is decreasing, and every strict change removes at least one grid point from a state value set. Thus the procedure reaches a fixed point $\xcalV^\star$ within $O(|S|N_\xi)$ iterations. With $\xi=(1-\gamma)\min\{\epsilon,\delta/2\}$ and fixed $\gamma$, $N_\xi$ is polynomial in $1/\epsilon$ and $1/\delta$, establishing the first claim.

If $\xcalV^\star(s_0)=\emptyset$, then $\calV^\star(s_0)=\emptyset$ by \Cref{lmm:nested-x}, so no exactly feasible policy exists. Otherwise, select a target at the initial state by solving
\[
    \vv^\star
    \in
    \argmax_{\vv \in \xcalV^\star(s_0)} v_1.
\]
Since $\xcalV^\star(s_0)$ is a convex polytope described as the convex hull of finitely many grid points (by construction of $\xPhi$), the above maximization reduces to a linear program over convex combination weights, and can therefore be solved in polynomial time.

By \Cref{lmm:eps-delta-achievable}, the vector $\vv^\star$ is $(\epsilon,\delta)$-achievable from $s_0$.
Let $\pi$ be the policy constructed in the proof of \Cref{lmm:eps-delta-achievable}, using a fixed deterministic rule for selecting a basic feasible LP solution at each expansion. Since $\calV^\star\subseteq\xcalV^\star$, the reward component of $\vv^\star$ is at least that of every exactly feasible policy. Hence $\pi$ is $(\epsilon,\delta)$-optimal.
We compute its action distributions using $\ActProbs_\xi$, which replaces $\Expand$ in $\ActProbs$ (\Cref{alg:act-probs}) with $\Expand_\xi$ (\Cref{alg:expand-x}).

The procedure takes $(\vv^\star,h)$ as input and returns the action distribution $\pi(h)$ prescribed at history $h$.
At each recursive step on a reachable history, the current target satisfies $\vv \in \xcalV^\star(s)$, so a valid output of $\Expand_\xi(\vv,s)$ can be found by solving an LP of polynomial size.

To bound the bit lengths through these recursive calls, write the target coordinates over a common denominator, using at most $L$ bits for each numerator and the denominator. The LP coefficient matrix is fixed by the instance and grid, and its right-hand side depends affinely on the target. By Cramer's rule, the variables of a basic feasible solution in standard form can therefore be written over a common denominator using at most $L+C$ bits per numerator and denominator, where $C$ is polynomial in the input size and $N_\xi$. This common denominator cancels when recovering $\ww$ by dividing a weighted sum of the $z$-variables by $\barpi(a)$. Since the grid coordinates are fixed, the next target also requires at most $L+C'$ bits per numerator and common denominator, for a polynomial bound $C'$ independent of the current target. Thus each expansion adds only polynomially many bits, so both target and LP solution bit lengths remain polynomial after at most $|h|$ expansions.

Using a fixed deterministic polynomial-time LP algorithm to select basic feasible solutions, the procedure performs at most one expansion per state in the input history and returns an arbitrary fixed distribution if the history is unreachable. Its total running time is thus polynomial in the stated quantities and $|h|$.

This yields an $(\epsilon,\delta)$-optimal policy $\pi$ whose action distribution $\pi(h)$ can be computed in polynomial time for any given history $h$.
\end{proof}

\end{document}